\documentclass[square,comma]{article} 
\usepackage[preprint]{neurips_2025}
\usepackage{hyperref}
\usepackage{url}
\usepackage{amssymb,amsmath,amsfonts}
\usepackage{bm}
\usepackage{algorithm}
\usepackage{algpseudocode}
\usepackage{color}
\usepackage{enumitem}
\usepackage{array}
\usepackage{booktabs}
\usepackage{multirow}
\usepackage{makecell}
\usepackage{subfig}
\usepackage{wrapfig}
\usepackage{float}
\usepackage{graphicx}
\graphicspath{{./figures/}}
\usepackage{natbib}
\setcitestyle{numbers,square}

\usepackage[utf8]{inputenc} 
\usepackage[T1]{fontenc}    
\usepackage{hyperref}       
\usepackage{url}            
\usepackage{booktabs}       
\usepackage{amsfonts}       
\usepackage{nicefrac}       
\usepackage{microtype}      
\usepackage{xcolor}         

\newtheorem{theorem}{Theorem}[section]
\newtheorem{proposition}[theorem]{Proposition}

\title{Proactive Constrained Policy Optimization with Preemptive Penalty}

\author{
Ning Yang\textsuperscript{1}\thanks{Correspondence to: Ning Yang <ning.yang@ia.ac.cn>}~~~
Pengyu Wang\textsuperscript{1}~~~
Guoqing Liu\textsuperscript{2}~~~
Haifeng Zhang\textsuperscript{1}~~~
Pin Lv\textsuperscript{1}~~~
Jun Wang\textsuperscript{3}~~~\\
\textsuperscript{1}Institute of Automation, Chinese Academy of Sciences~~~\\
\textsuperscript{2}Microsoft Research AI4Science~~~
\textsuperscript{3}University College London~~~
}

\begin{document}

\maketitle

\begin{abstract}

Safe Reinforcement Learning (RL) often faces significant issues such as constraint violations and instability, necessitating the use of constrained policy optimization, which seeks optimal policies while ensuring adherence to specific constraints like safety. Typically, constrained optimization problems are addressed by the Lagrangian method, a post-violation remedial approach that may result in oscillations and overshoots. Motivated by this, we propose a novel method named Proactive Constrained Policy Optimization (PCPO) that incorporates a preemptive penalty mechanism. This mechanism integrates barrier items into the objective function as the policy nears the boundary, imposing a cost. Meanwhile, we introduce a constraint-aware intrinsic reward to guide boundary-aware exploration, which is activated only when the policy approaches the constraint boundary. We establish theoretical upper and lower bounds for the duality gap and the performance of the PCPO update, shedding light on the method's convergence characteristics. Additionally, to enhance the optimization performance, we adopt a policy iteration approach. An interesting finding is that PCPO demonstrates significant stability in experiments. Experimental results indicate that the PCPO framework provides a robust solution for policy optimization under constraints, with important implications for future research and practical applications.
\end{abstract}

\section{Introduction}
\label{Introduction}

Security learning plays a crucial role in the field of computer science, as it is essential for ensuring data privacy, withstanding adversarial attacks, enhancing the security of intelligent systems, and meeting compliance requirements. Recently, this field has seen substantial advancements in tackling complex challenges such as safe exploration \cite{dalal2018safe}, the application of Lyapunov methods \cite{chow2018lyapunov, hao2024lyapunov}, and constrained optimization \cite{tessler2018reward, gu2024review, zhang2024cvar}. Despite these advancements, the dynamic nature of network security threats, coupled with the vast scale and complexity of computer systems, continues to pose significant challenges.

\begin{figure}[t]
    \vskip 0.2in
    \begin{center}    
        \subfloat[Lagrange multiplier]{
            \centering
            \includegraphics[width=0.3\linewidth]{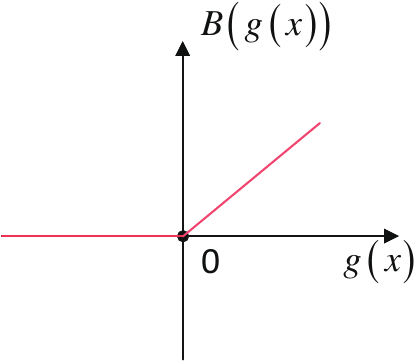} 
            \label{fig:Lagrange}
        }
        \hspace{0.05\textwidth}  
        \subfloat[Log-barrier extension]{
            \centering
            \includegraphics[width=0.3\linewidth]{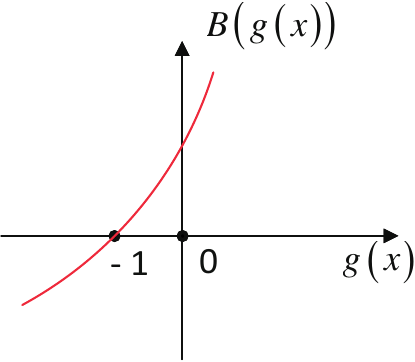} 
            \label{fig:Barrier}
        }
        \caption{Lagrange and barrier terms vs. constraint violations}
        \label{fig:LagrangeAndBarrier}
    \end{center}
\end{figure}

In the realm of safe Reinforcement Learning (RL), also known as constrained RL, the Lagrangian  method has emerged as a predominant tool \cite{ghosh2025safe}. Its simplicity and effectiveness in achieving optimal constraint satisfaction \cite{chow2019lyapunov, tessler2018reward}.
Nonetheless, this approach faces several challenges that hinder its application in constrained optimization problems. One issue is that the control of Lagrange multipliers exhibits hysteresis, implying that when the constraints are violated, the Lagrange multipliers cannot instantaneously transition to an expected value  \cite{stooke2020responsive, Honari2024safe}. Another challenge is that the Lagrangian method cannot serve as a barrier near the boundary of the constraints. In other words, when a policy approaches the constraint boundary, there is no penalty or change in gradient. This limitation of the method is demonstrated in Figure \ref{fig:LagrangeAndBarrier}, which illustrates the relationship between constraint violation value $g(x)$ and Lagrange multiplier or barrier items $B(g(x))$. Figure \ref{fig:Lagrange} uses the Lagrange multiplier method to demonstrate that the penalty value $B(g(x)) = 0$ when $g(x) \leq 0$ (constraint satisfied); when $g(x) > 0$ (constraint violated), the penalty value $B(g(x)) > 0$. Figure \ref{fig:Barrier} describes the extended barrier function method, where for $-1 < g(x) < 0$, the barrier term $B(g(x)) > 0$, demonstrating the \textit{preemptive penalty} mechanism adopted by the barrier function method.

Considering the intricate nature of constraint management in optimization problems and the shortcomings of traditional Lagrangian methods, we introduce a preemptive penalty to mitigate constraint violations and enhance optimization performance. We propose the Proactive Constrained Policy Optimization (PCPO) with a preemptive penalty. The advantages of our method are as follows: i) It exhibits a strictly positive gradient, which increases as a satisfied constraint approaches violation during optimization. This pushes the constraint back towards the achievable range. ii) Another crucial advantage is that the derivatives of our method provide the implicit dual variables, ensuring duality-gap guarantees. Our contributions in this paper are outlined as follows:

\begin{itemize}

\item \textit{Methodology}:
Within the PCPO approach, we address constraint violations by integrating a preemptive penalty, embedding the barrier term directly into the objective function. This approach proactively penalizes potential constraint breaches, circumventing the issue of zero gradients. Additionally, we introduce a constraint-aware intrinsic reward to guide boundary-aware exploration, capturing the potential impact degree of each sample on the strategy relative to the constraint boundary.

\item \textit{Theoretical Framework}:
Our analysis includes a derivation of approximate Kullback-Leibler (KL) divergence for the Gaussian model. We establish a theoretical upper bound on the duality gap, alongside a lower bound on the performance improvements following a PCPO update, ensuring a robust and stable theoretical foundation for our methodologies.

\item \textit{Experiment Results}:
Our results validate that the PCPO method effectively elevates the quality of solutions and markedly diminishes the incidence of constraint violations. Furthermore, the proposed method has been verified in terms of generalization and sensitivity analysis.

\end{itemize}

\section{Related Work}\label{Related Work}
Safe RL is a crucial branch of RL aimed at addressing constrained optimization or control problems, incorporating a variety of specialized methods to ensure safety during learning \cite{gu2024review}. Primal methods directly tackled the primal problem by alternating optimizing between the objective function and constraints, which could lead to instability in network training \cite{wachi2024safe}. The safeguard/safety layer approach, which included the projection method \cite{yang2020projection}, may face challenges in finding suitable projections. Penalty methods attempted to manage constraints through reward shaping or regularization, and choosing the right penalty factor remained difficult \cite{tessler2018reward}. Direct policy optimization leveraged a surrogate function, but this approach was susceptible to approximation errors \cite{chow2019lyapunov}.

Primal-dual methods, widely utilized in safe RL, are extensively developed to approximate the primal problem by optimizing the dual problem. However, Chow \textit{et al.} highlighted that updates to dual variables do not ensure immediate constraint satisfaction  \cite{chow2018risk}. Enhancing this approach, Achiam \textit{et al.} reset dual variables in each iteration to ensure consistent adherence to constraints \cite{achiam2017cpo}. 
Wagener \textit{et al.} adopted Lagrangian dual gradient ascent for updating Lagrange multipliers, thus enhancing model adaptability \cite{wagener2021safe}. Ding \textit{et al.}  proposed a periodically restarted optimistic primal-dual Proximal Policy Optimization (PPO) to effectively address time-varying constraints \cite{ding2023provably}. 
Ying \textit{et al.} investigated scalable primal-dual methods in safe multi-agent RL, broadening the applicability of these techniques to more complex systems \cite{ying2024scalable}. 
Expanding the scope,  Chen \textit{et al.} proposed an adaptive primal-dual method, attempting to solve the dual problem in safe RL by adjusting two adaptive learning processes to Lagrange multipliers \cite{chen2024safe}. 

Meanwhile, constrained intrinsic rewards in safe RL have also garnered extensive research attention \cite{gu2024balance}. To address the issue of introducing safety constraints, Kwon \textit{et al.} proposed a constrained reward framework to balance the performance objective of the total reward and the safety constraints \cite{kwon2024safe}. Furthermore, Li \textit{et al.} uses the metric distance between the current optimal policy and the theoretically optimal policy as an intrinsic reward, and optimizes the safety strategy through reward shaping \cite{li2025safe}.
However, despite these advancements, the typical application of penalties post-violation in these methods could still lead to potential network instability \cite{sinan2024ieee, su2025review}, prompting further exploration into preemptive penalty mechanisms to enhance safety measures.
\section{Constrained Markov Decision Process}
\label{Constrained Markov Decision Process}

A Constrained Markov Decision Process (CMDP) \cite{altman2021constrained} is denoted by a tuple $ \langle \mathcal S, \mathcal A, P,\nu, R, C \rangle$, where $\mathcal S$ represents the state space and $\mathcal A$ denotes the action space, ${P: \mathcal S \times \mathcal A \to \mathcal S} $ is a probabilistic transition process with components ${P_\pi }(s'|s) \in {\mathbb{R}}$, $\nu :\mathcal S \to [0, 1]$ is a probability distribution over initial states, ${R: \mathcal S \times \mathcal A \times \mathcal S} \to\mathbb{R}$ is the reward function, and set $C$ contains a set of cost functions ${C_i: \mathcal S \times \mathcal A \times \mathcal S} \to\mathbb{R}$, $i = 1, 2, ...,m$. A policy $\pi$  is a mapping from states to distributions over actions $ \mathcal{S} \times \mathcal{A} \to [0,1]$, and $\Pi$ is a set of policies. Our goal is to discover a stationary policy that maximizes the expected discounted reward $J(\pi ) := {\mathbb{E}_\pi } \left[ \sum\nolimits_{t = 0}^\infty {{\gamma ^t}R({s_t},{a_t})}\right]$ under constraints, where the agent's actions dictate the subsequent rewards and states. Considering the initial state ${s_0}$, the state-value function is $
{V^\pi }(s) := {\mathbb{E}_\pi } \left[ \sum\nolimits_{t = 0}^\infty  {{\gamma ^t}} R({s_t},{a_t})|{s_0} = s \right]$.
The action-value function \cite{DavidLecture} can similarly be decomposed $
{Q^\pi }(s,a) := {\mathbb{E}_\pi }\left[ \sum\nolimits_{t = 0}^\infty  {{\gamma ^t}} R({s_t},{a_t})|{s_0} = s,{a_0} = a \right]$. The advantage function \cite{mnih2016asynchronous} is ${A^\pi }(s,a) := {Q^\pi }(s,a) - {V^\pi }(s)$.
The discounted probabilistic transition distribution \cite{achiam2017cpo} is denoted by ${d_\pi }(s) := (1 - \gamma )\sum\nolimits_{t = 0}^\infty  {{\gamma ^t}} p_\pi ^t(s)$ where $p_\pi ^t(s) = P({s_t} = s|\pi )$ is a vector with $p_\pi ^t(s) \in {\mathbb{R}}$. The $C_i$-return, defined as the expected cumulative discounted cost for cost function \( i \), is given by 
${J_{{C_i}}}(\pi ) := {\mathbb{E}_\pi }[\sum\nolimits_{t = 0}^\infty  {{\gamma ^t}C_i({s_t},{a_t})} ]$
where \( C_i(s_t, a_t) \) represents the cost incurred at state \( s_t \) and action \( a_t \) under cost function \( i \). The set of feasible policies is ${{\Pi} _C} := \{ \pi  \in \Pi :{J_{C_i}}(\pi ) \le {d_i},i = 1,2,....,m\}$. The optimal policy of the CMDP is given by $\pi ^* = \arg {\max _{\pi  \in {\Pi _C}}}J(\pi )$.

Similar to the value function $V^ \pi$, action-value function $Q^ \pi$, and advantage function $A^ \pi$ for reward, we denote the cost value function, cost action-value function, and cost advantage function as $V_{{C_i}}^\pi$, $Q_{{C_i}}^\pi$, and $A_{{C_i}}^\pi$.

\section{Proactive Constrained Policy Optimization}
Building on this foundation, the implementation of PCPO with preemptive penalty can be summarized in two steps:
\begin{enumerate}
    \item \textit{Design of Constraint-Aware Intrinsic Reward:} A constraint-aware intrinsic reward is introduced to guide boundary-aware exploration, which is activated only when the policy approaches the constraint boundary and encourages behaviors that remain within the feasible region.
    \item \textit{Optimization with Preemptive Penalty:} The PCPO finds the constrained updated policy with an extended barrier function method featuring a preemptive penalty mechanism that adopts a conservative policy iteration. 
\end{enumerate}

\subsection{Problem Formulation}
Initially, a constrained optimization problem is formulated. Samples are then drawn from the environment to update policies, considering a set of parameterized policy spaces ${\Pi _\theta }$ with a fixed neural network. At the $k$-th iteration, the policy ${\pi _{{\theta _k}}}$ is updated to ${\pi _{{\theta _{k+1}}}}$ under constraints. The constrained objective function is defined as follows:
\begin{subequations}\label{eq:CMDP Optimization problem}
\begin{align}
\mathop {\max }\limits_{{\pi _\theta } \in {\Pi _\theta }}  & {\quad \mathop  \mathbb{E}\limits_{ s \sim {{d_{{\pi _{{\theta _k}}}}}}\hfill\atop
 a\sim {\pi _\theta }\hfill} [{A^{{\pi _{{\theta _k}}}}}(s,a)]}
\label{eq:CMDP objective function}\\
{\rm s.t.}& \quad {{J_{{C_i}}}({\pi _{{\theta _k}}}) + \frac{1}{{1 - \gamma }}\mathop \mathbb{E} \limits_{ s \sim {{d_{{\pi _{{\theta _k}}}}}}\hfill\atop
 a\sim {\pi _\theta }\hfill} \left[ A_{{C_i}}^{\pi_{\theta _k}} (s,a) \right] \le d_i }, \nonumber\\
& \quad i = 1,2,\ldots,m \label{eq:CMDP constraints1}\\
&{\mathop \mathbb{E} \limits_{s \sim {{d_{{\pi _{{\theta _k}}}}}}} [{D_{KL}}({\pi _\theta(.|s) }||{\pi _{{\theta _k}}}(.|s))] \le \delta },
\label{eq:CMDP constraints2}
\end{align}
\end{subequations}
where $\delta$ is a small constant. The objective function, as defined in \eqref{eq:CMDP objective function}, incorporates the advantage function in its expectation form, which provides a more accurate estimate of actions than the Q-function. Constraint \eqref{eq:CMDP constraints1} ensures compliance with the necessary conditions, while Constraint \eqref{eq:CMDP constraints2}, a trust region constraint, guarantees monotonic performance improvements by positing the presence of a non-negative expected advantage at each state. 

\subsection{Optimization with Preemptive Penalty Method}

A policy is employed in the previous subsection to enhance the quality of optimization. Building on this approach, this subsection focuses on reducing constraint violations through the use of constrained policy optimization, a type of policy search algorithm designed for CMDP. This algorithm includes updates that approximately solve constrained optimization problems, effectively minimizing potential penalties. Further drawing from the concept of barrier functions, the preemptive penalty method is employed to ensure that solutions remain within the feasible region. We propose the PCPO with a preemptive penalty by embedding barrier terms into the objective function \cite{kervadec2022constrained}. As the policy nears the boundary of the constraints, the preemptive penalty mechanism automatically activates, effectively reducing violations by proactively preventing them before they occur. The mathematical formulations for the simplified objective function \eqref{eq:CMDP objective function} and Constraints \eqref{eq:CMDP constraints1} are as follows: 
\begin{equation}
\begin{aligned}
f(\pi_{\theta} ) =\!\!\!\!\! \mathop  \mathbb{E}\limits_{ s \sim {{d_{{\pi _{{\theta _k}}}}}}\hfill\atop
 a\sim {\pi _\theta }\hfill} [{A^{{\pi _{{\theta _k}}}}}(s,a)]
\label{eq:Simplified objective function},
\end{aligned} 
\end{equation}
\begin{equation}
\begin{aligned}
{g_{{C_i}}}(\pi_{\theta} ) & \!\!=\!\! {{J_{{C_i}}}({\pi _{{\theta _k}}}) \!\!+ \!\!\frac{1}{{1 - \gamma }}\!\!\mathop \mathbb{E} \limits_{ s \sim {{d_{{\pi _{{\theta _k}}}}}}\hfill\atop a\sim {\pi _\theta }\hfill} \!\!\left[ A_{{C_i}}^{\pi_{\theta _k}} (s,a) \right] \!\!- \!\!d_i}\le 0 
\label{eq:Simplified constraints1}.
\end{aligned}
\end{equation}
To solve this constrained problem, we use a barrier function approach. Though this is not the standard Lagrangian dual formulation, it can be interpreted as an approximation that implicitly defines dual variables. We formulate the barrier problem as follows:
\begin{subequations}\label{eq:original Objective function with barrier problem}
\begin{align}
\mathop {\max }\limits_{{\pi_{\theta}  } \in {\Pi_{\theta} }}  & {\quad \bar{G}(\pi_{\theta} )}
\label{eq:original Objective function with barrier function}\\
{\rm s.t.}& \quad {\mathop \mathbb{E} \limits_{s \sim {{d_{{\pi _{{\theta _k}}}}}}} [{D_{KL}}(\pi_\theta||{\pi _{{\theta _k}}})] - \delta \le 0}, 
\end{align}\label{eq:original constraints1 function with barrier}
\end{subequations}
where $\bar{G}(\pi_{\theta} ) = f(\pi_{\theta} ) - \sum\limits_{i = 1}^m {{\varphi _\tau }({g_{{C_i}}}(\pi_{\theta} ))}$.
Here, $\varphi_\tau(.)$ is an extended log-barrier function that is convex and continuous:
\begin{equation}
\begin{aligned}
& \varphi_\tau({g_{{C_i}}}(\pi_{\theta}))= \begin{cases}-\frac{1}{\tau} \log (-{g_{{C_i}}}(\pi_{\theta})) & {g_{{C_i}}}(\pi_{\theta}) \leqslant-\frac{1}{\tau^2}, \\ \tau {g_{C_i}}(\pi_{\theta}) -\frac{1}{\tau} \log (\frac{1}{\tau^2}) + \frac{1}{\tau}& \text { otherwise. }\end{cases}
\end{aligned} \label{eq:continous barrier function}
\end{equation}
where $\tau$ is a given constant of the logarithmic barrier function. The first line of Eq.(\ref{eq:continous barrier function}) aligns with the standard log-barrier. To maintain continuity at the critical point $g_{{C_i}}(\pi_{\theta}) = -\frac{1}{\tau^2}$, the second line of the function in Eq.(\ref{eq:continous barrier function}) is specifically constructed, ensuring both mathematical and computational efficiency. And the gradient of $\varphi_\tau({g_{{C_i}}}(\pi_{\theta}))$ is:
\begin{equation}\label{nablaphi}
    \small \begin{aligned}
  &\nabla \varphi_\tau(g_{C_i}(\pi_{\theta})) = \frac{1}{\tau (\gamma-1) \cdot g_{C_i}} \mathbb{E}_{\substack{s \sim d_{\pi_{\theta_k}} \\ a \sim \pi_{\theta}}} \left[ \nabla_{\theta} \log \pi(a \mid s) \cdot \frac{A_{C_i}^{\pi_{\theta_k}}(s, a)}{(\gamma-1) \cdot\varphi_r(g_{C_i})} \right]
    \end{aligned}
\end{equation}
Note that problem \eqref{eq:original Objective function with barrier problem} can be viewed as an approximation to the dual problem of the primal problem \eqref{eq:CMDP Optimization problem}. The implicit dual variables $\lambda_i^*$ in this approximation are related to the barrier function derivatives, as will be shown in the duality gap theorem.

While the initial formulation of \(\bar G(\pi_{\theta})\) in problem \eqref{eq:original Objective function with barrier problem} integrates the objective function and barrier terms to address constraint violations proactively, it lacks an explicit mechanism to guide exploration behaviors, particularly when the policy is near constraint boundaries.

To mitigate this, we introduce the constraint-aware intrinsic reward \(I^{\pi_\theta}\) in problem \eqref{eq:original Objective function with barrier problem}, which can be constructed as a cost-sensitive approximation of the per-sample gradient contribution to the logbarrier term. From an information-theoretic perspective, this term captures how much each sample potentially shifts the policy with respect to the constraint boundary. 
The constraint-aware intrinsic reward $I^{\pi_\theta}_{C_i}$ is formulated proportionally to the magnitude of the cost advantage function normalized by the barrier term, reflecting the alignment between policy gradients and constraint violation risks. Specifically, the reward $I^{\pi_\theta}$  takes the following form:
\begin{equation}\label{correlated}
    I^{\pi_\theta}_{C_i} \propto \left|\frac{A^{\pi_\theta}_{C_i}(s,a)}{(1-\gamma) \cdot g_{C_i}(\pi_\theta)}\right|
\end{equation} 
To modulate the total exploration intensity during training, the intrinsic reward structure employs a gating mechanism based on proximity to constraint values, i.e., we multiply the normalized intrinsic reward by a gating function $\sigma(\alpha(\delta +g_{C_i}(\pi_\theta)))$, which activates only when the policy approaches the constraint boundary. Meanwhile, a Softmax function is used to map the scale to $(0, 1]$ for numerical stability:
\begin{equation}
    \begin{aligned}
        I^{\pi_\theta}_{C_i} =&  \sigma(\alpha(\delta + g_{C_i}(\pi_{\theta}))) \cdot \text{Softmax}\left(\beta \cdot \left|\frac{A_{C_i}^{\pi_{\theta}}(s, a)}{(1-\gamma) \cdot \max(-g_{C_i}(\pi_{\theta}), \epsilon)}\right|\right)
    \end{aligned}
\end{equation}
Here, $\sigma$ is the sigmoid function. This formulation resembles the ``$\text{score} \times \text{influence}$" structure of Fisher information, providing a theoretically grounded measure for boundary-level exploration value. The adjusted magnitude controls the relative incentive among eligible actions. This structure quantifies the sensitivity of model outputs to parameter changes. Change the triggering condition of the gating function to $\delta + g_{C_i}(\pi_{\theta}) \ge 0$. We only activate this intrinsic bonus when the policy direction reduces cost. It promotes behaviors that are more likely to remain within the feasible region and improve constraint satisfaction over time, without rewarding unsafe deviations. 

With this intrinsic reward, we can now reformulate the barrier problem to better balance reward maximization, constraint satisfaction, and safe exploration. The barrier problem is as follows:
\begin{subequations}\label{eq:Objective function with barrier problem}
\begin{align}
\mathop {\max }\limits_{{\pi_{\theta}  } \in {\Pi_{\theta} }}  & {\quad {G}(\pi_{\theta} )}
\label{eq:Objective function with barrier function}\\
{\rm s.t.}& \quad {\mathop \mathbb{E} \limits_{s \sim {{d_{{\pi _{{\theta _k}}}}}}} [{D_{KL}}(\pi_\theta||{\pi _{{\theta _k}}})] - \delta \le 0}, 
\end{align}\label{eq:constraints1 function with barrier}
\end{subequations}
where ${G}(\pi_{\theta} ) = f(\pi_{\theta} ) - \sum\limits_{i = 1}^m {{\varphi _\tau }({g_{{C_i}}}(\pi_{\theta} ))}+\eta \sum\limits_{i = 1}^m I^{\pi_\theta}_{C_i}$. Here, $\eta = \cfrac{\omega \cdot G^{\max}}{I_{C_i}^{\max}+\epsilon} > 0$, and $\epsilon$ is an extremely small constant to avoid the denominator being zero. To further validate the effectiveness of this intrinsic reward in enhancing optimization performance, we establish the following proposition. Proposition \ref{proposition1} shows that intrinsic reward can provide an additional positive boost for policy optimization. The proof is provided in the Appendix.
\begin{proposition}[Enhancement with Intrinsic Reward]\label{proposition1}
    Let $\pi_{k+1}$ and $\bar{\pi}_{k+1}$ be the policies updated from the same $\pi_k$ under the PCPO objective with and without the constraint intrinsic reward $I^{\pi_\theta}_{C_i}$, respectively. Define the objective function without the intrinsic reward as $\bar G(\pi_k)$. Then, we have:
\begin{equation}
\begin{aligned}
    G(\pi_{k+1}) - G(\pi_k) \ge & [\bar G({\pi}_{k+1}) - \bar G(\pi_k)] + \eta \sum \limits_{i=1}^{m} \left( I^{\pi_{k+1}}_{C_i} - I^{\pi_k}_{C_i} \right).
\end{aligned}
    \end{equation}
\end{proposition}

\subsection{Theoretical Analysis of Preemptive Penalty}
Next, our method is specifically designed to address CMDP problems using a conservative policy iteration scheme. Throughout the iteration optimization process, our proposed method adheres to the primal constraints. To enhance the optimization of dual problems \eqref{eq:Objective function with barrier problem}, an upper bound of the duality gap and dual variables are derived as follows:

\begin{theorem}[Duality Gap]
An upper bound of the duality gap between primal problem \eqref{eq:CMDP Optimization problem}  and dual problem \eqref{eq:Objective function with barrier problem} is given by
\begin{equation}
\begin{aligned}
G\left(\boldsymbol{\lambda}^*\right)-J\left(\boldsymbol \pi^*\right) \leq \frac{m}{\tau}+\eta \sum\limits_{i=0}^{m}I^{\max}_{C_i},
\label{eq:(9)}
\end{aligned}
\end{equation}
where $\boldsymbol{\pi}^*$ is the optimal policy, $m$ is the number of constraints and \(I^{\text{max}}\) is the upper bound of the expected intrinsic reward. The optimal implicit Lagrangian dual variables satisfy 
\begin{equation}
\begin{aligned}
& \lambda_i^*=\left\{\begin{array}{cc}
-\frac{1}{\tau g_{{C_i}}\left(\boldsymbol {\pi^*}\right)} &  g_{{C_i}}\left(\boldsymbol \pi^*\right) \leq-\frac{1}{\tau^2}, \\
\tau & \rm{ otherwise. }
\label{eq:(10)}
\end{array}\right. \\
\end{aligned}
\end{equation}
\label{Duality Gap}
\end{theorem}
For convenience, to analyze the optimal implicit Lagrangian dual variables, we only rewrite \( G\left(\boldsymbol{\pi}^*\right) \) as \( G\left(\boldsymbol{\pi}^*, \boldsymbol{\lambda}^*\right) \) in Theorem \ref{Duality Gap}. Theorem \ref{Duality Gap} describes the relationship between the duality gap and the parameter $\tau$. As $\tau$ increases that leads to the gap diminishes. This upper bound shows how dual variables are adjusted in response to violations. The proof process is elaborated in Appendix \ref{Proof of Theorem Duality Gap}.

In the PCPO algorithm, we denote the policy at the $k$-th iteration as $\pi_{\theta_k}$, and the updated policy as $\pi_{\theta_{k+1}}$. The update rule of the PCPO algorithm refers to the mathematical mechanism for transitioning from $\pi_{\theta_k}$ to $\pi_{\theta_{k+1}}$, which is achieved by solving Eq.(\ref{eq:Objective function with barrier problem}), maximizing ${G}(\pi_{\theta})$ while satisfying the KL-divergence constraint to obtain $\pi_{\theta_{k+1}}$. This update rule ensures that the transition from $\pi_{\theta_k}$ to $\pi_{\theta_{k+1}}$ both improves performance and maintains constraint satisfaction. An explicit lower bound is provided for performance improvements between two adjacent iterations in PCPO, which provides a benchmark to evaluate the effectiveness of the PCPO method. The theoretical result is represented as follows:
\begin{theorem}[PCPO Update Performance]
A lower bound for consecutive policies $\pi_{k+1}$ and $\pi_k$ improvements is
\begin{equation}\label{eq:PCPO Update Performance}
\small\begin{aligned}
&G({\pi _{k + 1}}) - G({\pi _k}) \geqslant\\
& \left\{ {\begin{array}{*{20}{c}}
{{\eta }({\pi _{k + 1}})- \frac{1}{\tau }\sum\limits_{i = 1}^m {\log \left( {2 - \frac{{{d_i}}}{{2{\eta _{{C_i}}}({\pi _{k + 1}})}}} \right)\!\!-\!\!\eta\!\sum\limits_{i = 1}^m\!\!  I^{\max}_{C_i}} }, {g_{\!{C_i}\!}}(\pi_k) \!\leqslant\!\!-\!\frac{1}{\tau^2},\\
{ \!- m{d_i}+ \!\!{\eta}({\pi _{k + 1}})+\sum\limits_{i = 1}^m {\!\!( {\eta _{{C_i}}}\!({\pi _k})\!\! +\! \!{\eta _{{C_i}}}({\pi _{k + 1}}))\!\!-\eta\sum\limits_{i = 1}^m \!\!I^{\max}_{C_i}}},{{\rm{ otherwise. }}}
\end{array}} \right.
\end{aligned}
\end{equation}
where 
${\epsilon^{{\pi _{k + 1}}}}={\max _s} {\left|   {{\mathbb{E}_{a\sim{\pi _{k + 1}}}}\left[ {{A^{{\pi _k}}}} \right]} \right|}$, $\quad \epsilon_{C_i}^{\pi_k}={\max _s} {\left|   {{\mathbb{E}_{a\sim{\pi _{k}}}}\left[ {{A_{C_i}^{{\pi _k}}}} \right]} \right|}$, $\epsilon_{{C_i}}^{{\pi _{k + 1}}} = {\max _s}|{\mathbb{E}_{a\sim \pi_{k + 1}}}[A_{{C_i}}^{{\pi _{k + 1}}}]|$,

${\eta}({\pi _{k + 1}}) = \frac{{ - \sqrt {2\delta } \gamma \epsilon^{{\pi _{k + 1}}}}}{{{{(1 - \gamma )}^2}}}$,${\eta _{{C_i}}}({\pi _k}) = \frac{{ - \sqrt {2\delta } \gamma \epsilon_{{C_i}}^{{\pi _{k }}}}}{{{{(1 - \gamma )}^2}}}$, and ${\eta _{{C_i}}}({\pi _{k + 1}}) = \frac{{ - \sqrt {2\delta } \gamma \epsilon_{{C_i}}^{{\pi _{k + 1}}}}}{{{{(1 - \gamma )}^2}}}$.
\label{PCPO Update Performance}
\end{theorem}
 
Theorem \ref{PCPO Update Performance} reflects the minimal improvement between two iterations, ensuring reward maximization and constraint satisfaction. This lower bound works as a crucial check that guides parameter tuning and enforces consistent policy evolution. The proof is provided in Appendix \ref{Proof of Theorem PCPO Update Performance}.



\begin{proposition}[Advantage in Cumulative Constraint Violations]
    In a CMDP with the constraint set $\{J_C(\pi) \leq d\}$, for any initial policy $\pi_0$ and number of iterations $T$, the cumulative constraint violation of the PCPO method (denoted $V_{\text{P}}(T)$) and that of Lagrangian-based safe RL methods (denoted $V_{\text{L}}(T)$) satisfy:
    \begin{equation}
        V_{\text{P}}(T) \leq V_{\text{L}}(T) - \Delta(T),
    \end{equation}
where $\Delta(T) > 0$ is a gap term positively correlated with $T$, and $\Delta(T) \to \infty$ as $T \to \infty$.
\end{proposition}

Based on Theorem \ref{PCPO Update Performance}, we further compare the improvement between adjacent iterations during the training processes of PCPO and Lagrangian-based safe RL methods in Proposition 4. Lagrangian-based safe RL methods effectively manage constraints using Lagrange multipliers to ensure safety while maximizing rewards, as demonstrated in \cite{chow2019lyapunov, tessler2018reward, stooke2020responsive}. The proof of Proposition 4 is provided in Appendix.

\section{Practical Implementation}
\label{Practical Implementation}
\subsection{Parameterized Objectives and Constraints} This section delves into the practical implementation where the dual problem, as outlined in Eq.(\ref{eq:Objective function with barrier problem}), is resolved directly despite its computational expense and inefficiency. The focus shifts to exploring parameterized policies to develop a practical algorithm capable of managing finite sample counts and arbitrary initial parameters. The notation from the previous sections has been simplified using the parameter vector $\theta$. Therefore, the parameterized objective function is subject to constraints and its gradients are 
\begin{equation}\label{eq:parameterized
objective function}
\begin{aligned}
\mathop {\max }\limits_\theta  G(\theta ) = \mathop {\max }\limits_\theta  f(\theta ) - \sum\limits_{i = 1}^m {{\varphi _\tau }({g_{{C_i}}}(\theta ))}+\eta \sum\limits_{i = 1}^m I^{\pi_\theta}_{C_i}(\theta), 
\end{aligned}
\end{equation}
\begin{equation}\label{eq:the gradient of the parameterized objective function}
\begin{aligned}
\nabla G(\theta )= \nabla f(\theta ) - \sum\limits_{i = 1}^m \nabla {\varphi _\tau }({g_C}_{_i}(\theta ))\nabla {g_C}_{_i}(\theta )+\eta \nabla \sum\limits_{i = 1}^m I^{\pi_\theta}_{C_i}(\theta).
\end{aligned}
\end{equation}
In addition, the Fisher Information Matrix (FIM) is employed, utilizing analytical computations of the Hessian matrix derived from the KL divergence. As parameterized policies are considered, the previous notation will be extended to incorporate functions of $\theta$ instead of $\pi$.

The parameter update for Eq.\eqref{eq:the gradient of the parameterized objective function} is approximated employing a linear approximation of the objective function and a quadratic approximation for the KL divergence constraint.
\begin{subequations}\label{eq:parameter update problem}
\begin{align}
{\theta _{k + 1}} = \arg \mathop {\max }\limits_\theta  [{\nabla _\theta }{G^T}\left( \theta  \right){|_{\theta  = {\theta _k}}} \left( {\theta  - {\theta _k}} \right)]
\label{eq:parameter update objective},\\
 \text { s.t. } 
   \frac{1}{2}\left(\theta-\theta_k\right)^T H\left(\theta-\theta_k\right) \leq \delta, 
\end{align}\label{eq:parameter update constraints}
\end{subequations}
where $H$ is the Hessian matrix of the KL-divergence that is given by \begin{equation}\label{eq:Hessian matrix}
H\left(\theta_k\right)_{i j}\!\!=\!\!\left.\frac{\partial}{\partial \theta_i} \frac{\partial}{\partial \theta_j} \underset{s \sim d_{\pi_{\theta_k}}}{{\mathbb E}}\!\left[D_{KL}\left(\theta \| \theta_k\right)\right]\right|_{\theta=\theta_k}. 
\end{equation}

It is important to note that in practice, the Fisher information matrix $H$ may not always be invertible. To address this issue, we consider a regularized version of the Fisher information matrix $\hat{H} = H + \lambda I$, where $\lambda > 0$ is a small regularization parameter and $I$ is the identity matrix. This ensures $\hat{H}$ is positive definite and invertible. 

We consider a transformation $p = \hat{H}^{1/2}(\theta - \theta_k)$ in Eq.(\ref{eq:parameter update problem}), then Eq.(\ref{eq:parameter update problem}) is equivalent to optimizing a linear function on a norm ball:
\begin{subequations}\label{eq:transformed_problem}
\begin{align}
\max_p \quad & [\nabla_\theta G^T(\theta)|_{\theta = \theta_k} \hat{H}^{-1/2}p] \label{eq:transformed_objective}\\
\text{s.t.} \quad & \frac{1}{2}p^Tp \leq \delta \label{eq:transformed_constraint}
\end{align}
\end{subequations}

With this reformulation, the desired result follows in a relatively straightforward manner. The solution is simply:
\begin{equation}\label{eq:p_solution}
p^* = \sqrt{2\delta} \cdot \frac{\hat{H}^{-1/2}\nabla G}{\|\hat{H}^{-1/2}\nabla G\|_2}
\end{equation}
Transforming back to the original variable in Eq.(\ref{eq:parameter update problem}) is:

\begin{equation}\label{eq:theta in Dual function}
\theta^* = \theta_k + \sqrt{2\delta} \cdot \frac{\hat{H}^{-1}\nabla G}{\sqrt{\nabla G^T \hat{H}^{-1} \nabla G}}
\end{equation}

This approach does not require the strict feasible point assumption in Eq.(\ref{eq:parameter update problem}) and effectively handles cases where the Fisher information matrix may be non-invertible. 

\subsection{Estimation of Objectives and Constraints} Generalized Advantage Estimation (GAE) is leveraged to accurately estimate the policy gradient, providing a robust foundation for subsequent policy updates. In this process, a sequence of states is gathered through sampling and policy simulation over a specified number of timesteps to produce a trajectory. At each state-action pair, the Q-value is calculated by summing the future rewards discounted over the trajectory's length.

After estimating the policy gradient, we implement these calculations within a structured policy iteration framework. Algorithm 1 shows a policy iteration method based on the performance improvement bound in \eqref{eq:PCPO Update Performance}, employing the minimization-maximization algorithm \cite{Hunter2004MM}. A detailed pseudocode of the supplementary materials is available in the Appendix. 

\begin{algorithm}[tb]
\caption{PCPO}
\begin{algorithmic}[1]
\State Initialize policy network $\pi_0=\pi_{\theta_0}$
\For{$k = 0,1,2,...$}
    \State Run $\pi_k=\pi_{\theta_k}$ and store trajectories in $\mathcal{D}$
    \State Estimate $\nabla \hat G(\theta ), \nabla {\hat \varphi _\tau }({g_C}_{_i}(\theta ))\nabla {g_C}_{_i}(\theta ), \nabla I_{C_i}^{\pi_\theta}(\theta)$, $\hat H$ with $\mathcal{D}$
    \State Update $\theta_{k+1}$ using Eq.(\ref{eq:parameter update problem})
    \State Empty $\mathcal{D}$
    \EndFor 
    \end{algorithmic}
\end{algorithm}

\section{Experiments}
\label{Experiments} 

    \begin{figure}
    \vskip 0.2in
    \begin{center}    
    \begin{minipage}{\textwidth}  
    \centering{\includegraphics[width=1\linewidth, height=0.46\linewidth]{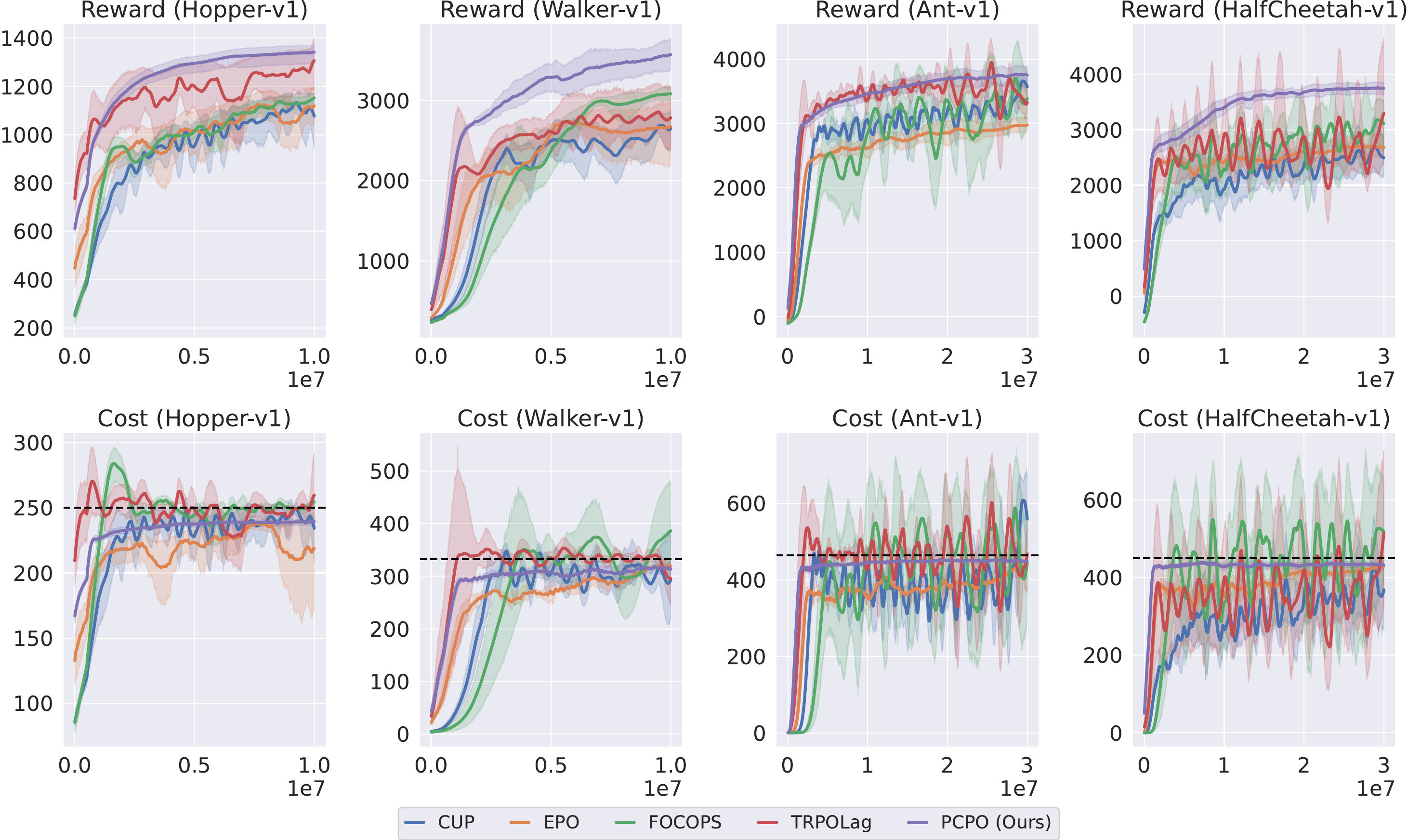}}
        \caption{The average performance for CUP, EPO, FOCOPS, TRPOLag and PCPO over 6 seeds (about 1000 bootstrap samples). The x-axis indicates the total number of samples, while the y-axis shows the mean total reward/cost return from the most recent 100 episodes. The shaded regions indicate the bootstrap normal 95\% confidence interval.}
        \label{fig1}
    \end{minipage}
    \end{center}
    \end{figure}
    \hfill

\begin{table}[t]

 \caption{Average episode return and cost for CUP, EPO, FOCOPS, TRPOLag and PCPO over 6 seeds. The results include the bootstrap mean and normal 95\% confidence interval, derived from 1000 bootstrap samples. Cost thresholds are indicated in brackets beneath the environment names.}\label{table1}
    \begin{minipage}{\textwidth}  
    \setlength{\tabcolsep}{1mm}
    \vskip 0.15in
    \hspace{-2cm}
    \begin{small}
    \begin{sc}
    \begin{tabular}{cccccccc}
        \toprule
        Environment & & CUP & EPO & FOCOPS & TRPOLag & PCPO \\
        \hline
        Hopper-v1 & Return & $1077.46\pm 175.83$ & $1119.73\pm105.76$ & $1152.3\pm28.28$& $1306.79\pm163.54$ & $\bm{1342.09\pm59.04}$ \\
        (250.) & Cost & $234.2\pm32.88$ & $218.66\pm62.16$ & $254.67\pm8.06$ & $259.77\pm40.05$ & $239.5\pm3.3$\\
        \hline
        Walker2d-v1 & Return & $2649.7\pm352.63$ & $2672.71\pm789,23$&$3084.23\pm126.58$& $2785.54\pm543.28$ & $\bm{3572.63\pm271.53}$ \\
        (333.) & Cost & $293.37\pm122.19$ & $318.81\pm12.32$& $386.46\pm117.8$& $295.79\pm61.1$ & \textbf{$313.31\pm18.01$}\\
        \hline
        Ant-v1   & Return & $3569.9\pm218.87$ &$2975.84\pm170.82$ & $3388.96\pm611.34$ & $3444.09\pm374.72$ & $\bm{3752.22\pm228.49}$ \\
        (465.) & Cost & $559.28\pm180.14$ & $441.87\pm19.45$ & $443.78\pm209.33$ & $466.57\pm109.82$ & $448.45\pm8.61$  \\
        \hline
        HalfCheetah-v1 & Return & $2494.21 \pm523.8$ & $2680.16\pm318.03$ & $3111.29\pm668.6$ & $3302.67\pm1633.98$ & $\bm{3746.79\pm167.41}$ \\
        (450.) & Cost & $369.16\pm128.1$ & $429.31\pm18.3$ & $518.85\pm188.42$ & $515.33\pm294.2$ & $431.79\pm15.94$ \\
        \bottomrule
   \end{tabular}
   \end{sc}
   \end{small}
   \end{minipage}
\end{table}

The efficacy of the PCPO algorithm is demonstrated with various robotic agents on both Safe Velocity and Safe Navigation tasks, which are integrated into the Safety-gymnasium \cite{ji2023safety} and executed using the MuJoCo physical simulator. Specifically, for the Safe Velocity tasks, four distinct robotic agents are trained to move either in a straight line or across a two-dimensional plane, while adhering stringently to imposed speed limits. Additionally, the circular motion mode is implemented within the Safe Navigation tasks, and two robotic agents are trained accordingly.

Five distinct RL algorithms are employed for comparative analysis within these environments. CUP \cite{yang2022cup} and FOCOPS \cite{zhang2020first} represent constrained policy update methods that explore the policy parameter space to satisfy constraints and maximize the objective function. TRPOLag \cite{schulman2015trust} combines the trust region method of TRPO with the Lagrangian multipliers approach to solve the max-min optimization problem.
EPO \cite{Gao2024ExteriorPP}, a variant of PPO, incorporates refined mechanisms into the policy update process,  enabling efficient constraint satisfaction and safe exploration. Further details are provided in the Appendix.

\subsection{Evaluation on Safe Velocity Task}

Four MuJoCo environments are selected for training a robotic agent to walk, subject to a speed limit. The cost thresholds are set at 50 percent of the cost required by an unconstrained TRPO agent, which is trained with 10 million samples. According to Figure \ref{fig1}, PCPO generally outperforms other algorithms in most tasks while adhering to the speed limits, surpassing CUP, EPO,  FOCOPS, and TRPOLag in every simulation. Furthermore, the integration of the preemptive penalty method into the PCPO algorithm demonstrates a certain capacity to handle constraint violations, exhibiting notable stability and constraint satisfaction abilities, such as the assumption that the initial policy is theoretically feasible. As training progresses, its cost gradually decreases and stabilizes, indicating a robust capacity for constraint adherence. Notably, while PCPO may not exhibit a significant reward advantage over TPROLag in Ant-v1, it substantially reduces the constraint violations and oscillations, resulting in a more stable and safety-compliant learning process. In Hopper-v1, TRPOLag exhibits a trend suggesting potential improvement with more iterations. However, we hypothesize that this may be partially due to TRPOLag’s greater exploration in unsafe regions before eventually converging to a higher-reward solution. Overall, our analysis confirms that PCPO effectively executes all tasks.
In Table \ref{table1}, we summarize the performance of all five algorithms.

\subsection{Evaluation on Safe Navigation Task}
For these tasks, we examine the performance of the point and car agents within the Safe Navigation tasks, particularly focusing on the circular motion mode. The objective is to train a robotic agent that can navigate within a circular perimeter while avoiding the boundaries of the environment. Consistent with prior tasks, we display the learning curves and provide numerical summaries. Notably, our findings reveal that PCPO effectively enforces the constraints in both experiments while achieving a commendable reward.

As depicted in Figure A1 and Table A3 in the Appendix, similar to the results of the experiment on Safe Velocity tasks, CUP, EPO, FOCOPS, and TRPOLag exhibit considerable fluctuations in meeting the target constraints, whereas our method remains highly stable and consistently falls below the cost limitation thresholds throughout the learning process. The result indicates that PCPO not only minimizes constraint violations but also converges to a superior solution.

\subsection{Generalization Analysis}
RL algorithm relies heavily on interaction with the environment, adjusting decision-making processes based on rewards and penalties. Traditional methods partition datasets into training, validation, and test sets to evaluate algorithm performance through simulated assessments, yet fail to adequately maintain consistency with RL training dynamics. Thus, such fixed dataset allocations are insufficient for extrapolating algorithm performance levels in actual tasks. In contrast, adaptation to complex tasks and dynamic changes can be more effectively achieved by interacting directly with the environment and learning through trial and error. This approach leads to more accurate solutions.

 To better assess the generalization of the PCPO algorithm, an approach utilizing fixed and unseen random seeds is implemented. This method evaluates performance under varied conditions, reflecting real-world scenarios more accurately. Experiments demonstrate that the PCPO algorithm outperforms others by effectively adapting to different test environments and consistently enhancing performance while maintaining lower costs. For specific experimental details and data, please refer to the Appendix.

 \subsection{Sensitivity Analysis}
In this analysis, it is investigated the dependency of the PCPO algorithm's performance on the hyperparameters $\tau$. The experiment investigations demonstrate that the algorithm exhibits robustness to variations in this parameter. Experiments are conducted on robots operating under speed constraints to validate this observation.

We conducted experiments with PCPO across 4 distinct $\tau$ values, simultaneously maintaining all other parameters fixed. Finally, the experiment observes that environments presenting greater difficulty, exhibit heightened sensitivity to parameter selections. Nevertheless, PCPO demonstrates substantial insensitivity to hyperparameter choices overall. Additionally, we note that our methods gain a more significant return and cost with a greater selection of $\tau$. We also investigate the impact of the weighting parameter $\omega$, which controls the relative contribution of the constraint-aware intrinsic reward with respect to the environment reward. See Appendix for details and data of specific results.

\subsection{Ablation Study}
The ablation studies highlight the effectiveness of preemptive penalties in the PCPO method, unlike the TRPOLag method, which does not utilize these penalties, as demonstrated in Figures \ref{fig1} and A1 in the Appendix, and Tables \ref{table1} and A3 in the Appendix. The importance of intrinsic reward for dynamically shaping exploration and improving the effectiveness of safe policy learning is emphasized. We also evaluate the effectiveness of our designed barrier function. The results of Safe Velocity tasks indicate significant performance improvements with the PCPO method. For further details, see Appendix.

\section{Conclusion}

In this article, we introduce a constrained optimization method PCPO, which integrates KL divergence, preemptive penalty methods, and a constraint-aware intrinsic reward to effectively address common issues such as oscillation, overshooting, and zero gradients. Experimental results demonstrate that compared to existing algorithms, our approach enhances safety during the learning process and exhibits superior performance across various standard testing environments, particularly in dynamic and complex uncertain settings. This method significantly reduces constraint violations during policy optimization, demonstrates strong adaptability to varying cost constraints, and possesses robust generalization capabilities. The implementation of PCPO is straightforward, making it accessible to researchers in other fields, especially in applications where safety is critically important. 

\newpage

\bibliographystyle{unsrt}

\newpage

\appendix
\setcounter{table}{0}   
\setcounter{figure}{0}
\renewcommand{\thetable}{A\arabic{table}}
\renewcommand{\thefigure}{A\arabic{figure}}
\section{Preliminaries}\label{Preliminaries}
Recalling the definitions for $p_\pi^t(s)$ and ${P_\pi}(s'|s)$, we can express their relationship as follows:
 \begin{equation}
\begin{array}{l}
p_\pi ^t(s') = \sum_{s} {P_\pi }(s'|s)p_\pi ^{t - 1}(s) = {P_\pi^t}\nu, 
\end{array} \label{eq:relationship for discounted transition distribution}
\end{equation}
where $\nu$ a probability distribution
over initial states. 
The discounted probabilistic transition distribution is denoted by 
\begin{equation}
\begin{array}{l}
{d_\pi }(s) = (1 - \gamma )\sum\limits_{t = 0}^\infty  {{\gamma ^t}} p_\pi ^t(s).
\end{array} \label{eq:discounted transition distribution}
\end{equation}
Combining \eqref{eq:discounted transition distribution} with \eqref{eq:relationship for discounted transition distribution}, the discounted transition distribution is rewritten as
\begin{equation}
\begin{array}{l}
{d_\pi }(s) = (1 - \gamma )\sum\limits_{t = 0}^\infty  {{\gamma ^t}{P_\pi^t}\nu }, 
\end{array} \label{eq:rewrite discounted transition distribution}
\end{equation}
for \eqref{eq:rewrite discounted transition distribution} which is reformulated by multiplying ${(I - \gamma {P_\pi})}$ on both sides, and then we have 
\begin{equation}
\begin{array}{l}
{d_\pi }(s) = (1 - \gamma )(I - \gamma P_\pi)^{-1}\nu.
\end{array} \label{eq:rewrite3 discounted transition distribution}
\end{equation}
Note that inserting the value function, the equality is obtained
\begin{equation}
\begin{array}{l}
(1 - \gamma )\mathop \mathbb{E}\limits_{s \sim \nu } [V^\pi(s)] = \mathop \mathbb{E}\limits_{s \sim {d_\pi }(s)} [r(s, \pi(s))]
\end{array} \label{eq:expected value function} 
\end{equation}
The discounted reward is written as \cite{achiam2017cpo}
\begin{equation}
J(\pi ) = \frac{1}{{1 - \gamma }}\mathbb{E}_{s \sim d_\pi, a \sim \pi}[r(s,a)].
\label{eq:discounted reward}
\end{equation}
Next, using \eqref{eq:rewrite3 discounted transition distribution}, the discounted reward \eqref{eq:discounted reward} can be expressed as
\begin{equation}
\begin{array}{l}
\begin{array}{l}
J(\pi ) \!\!= \!\!\!\mathop \mathbb{E}\limits_{s\sim \nu } [V^\pi(s)]\!\! + \!\!\frac{1}{{1 - \gamma }}\!\!\mathop \mathbb{E}\limits_{s\sim {d_\pi }(s)\hfill\atop
{a\sim \pi \hfill}} [{A^\pi }(s,a)]
\end{array}
\end{array} \label{eq:rewrite discounted reward}
\end{equation}
where ${A^\pi }(s,a) = {Q^\pi }(s,a) - V^\pi(s)$ is the advantage function. For similarity, the discounted accumulate cost using \eqref{eq:rewrite3 discounted transition distribution} is

\begin{equation}
\begin{array}{l}
J_{{C_i}}(\pi ) =  \mathop \mathbb{E}\limits_{s\sim \nu } [V_{{C_i}}^\pi(s)] + \frac{1}{{1 - \gamma }}\mathop \mathbb{E}\limits_{s\sim {d_\pi }(s)\hfill\atop
{a\sim \pi \hfill}} [{A_{{C_i}}^\pi }(s,a)].
\end{array} \label{eq:rewrite discounted cost}
\end{equation}

Based on the \eqref{eq:rewrite discounted reward} and \eqref{eq:rewrite discounted cost}, we can obtain the objective function \eqref{eq:CMDP Optimization problem} \cite{achiam2017cpo}.

\section{Proof of Proposition 1}
\textit{Proof:} Consider the PCPO objective function in the form:
\begin{equation}
    G(\pi) = f(\pi) - \sum_{i=1}^{m} \phi_\tau(g_{C_i}(\pi)) + \eta \sum_{i=1}^{m} I^{\pi}_{C_i},
\end{equation}
and its counterpart without the intrinsic reward:
\begin{equation}
    \bar{G}(\pi) = f(\pi) - \sum_{i=1}^{m} \phi_\tau(g_{C_i}(\pi)).
\end{equation}
At first, let $\bar\pi_k=\pi_k$. By the update rule, $\pi_{k+1}$ and $\bar{\pi}_{k+1}$ satisfy:
$\pi_{k+1} = \arg\max_{\pi} G(\pi)$ and $\bar{\pi}_{k+1} = \arg\max_{\pi} \bar{G}(\pi), $
with both under the same KL-divergence trust region constraint.

We first compute the difference in $G$ between consecutive iterations for \(\pi_{k+1}\):
\begin{equation}
\begin{aligned}
     &G(\pi_{k+1}) - G(\pi_k) \\=& [f(\pi_{k+1}) - f(\pi_k)] - \sum \limits_{i=1}^{m} [\phi_\tau(g_{C_i}(\pi_{k+1})) - \phi_\tau(g_{C_i}(\pi_k))]+ \eta \sum \limits_{i=1}^{m} [I^{\pi_{k+1}}_{C_i} - I^{\pi_k}_{C_i}],
\end{aligned}
   \end{equation}
Next, we compute the same difference for \(\bar{\pi}_{k+1}\):
\begin{equation}
\begin{aligned}
        &\bar{G}({\pi}_{k+1}) - \bar{G}({\pi}_k) = [f(\bar{\pi}_{k+1}) - f(\bar{\pi}_k)] - \sum \limits_{i=1}^{m} [\phi_\tau(g_{C_i}(\bar{\pi}_{k+1})) - \phi_\tau(g_{C_i}(\bar{\pi}_k))].
\end{aligned}
\end{equation}

Since $\pi_{k+1}$ is optimized under a richer objective (includes $I^{\pi}_{C_i}$), and the trust region ensures both updates are close, we can assume $f(\pi_{k+1}) \approx f(\bar{\pi}_{k+1})$.
Additionally, the intrinsic reward pushes the policy inward, meaning $g_{C_i}(\pi_{k+1}) \le g_{C_i}(\bar{\pi}_{k+1})$. This implies $\phi_\tau(g_{C_i}(\pi_{k+1})) \le \phi_\tau(g_{C_i}(\bar{\pi}_{k+1}))$. 

Thus, we obtain:
\begin{equation}
\begin{aligned}
    G(\pi_{k+1}) - G(\pi_k) \ge & [\bar G({\pi}_{k+1}) - \bar G(\pi_k)]  + \eta \sum \limits_{i=1}^{m} \left( I^{\pi_{k+1}}_{C_i} - I^{\pi_k}_{C_i} \right).
\end{aligned}
    \end{equation}



\section{Proof of Theorem 2}\label{Proof of Theorem Duality Gap}

\textit{Proof}:
Let $\boldsymbol \pi^*$ be the optimal policy of problem Eq.(1). 
When considering the revised barrier problem in Eq.(9) where \(G(\pi_{\theta}) = f(\pi_{\theta}) - \sum_{i=1}^{m} \varphi_{\tau}(g_{C_{i}}(\pi_{\theta})) + \eta \sum_{i=1}^{m}I^{\pi_{\theta}}_{C_i}\), we assume \(\pi^*\) approximately satisfies the optimality condition for the minimal value of Eq.(9).
Due to the inclusion of the constraint-aware intrinsic reward \(I^{\pi_{\theta}}_{C_i}\), the gradient condition now accounts for its contribution, leading to:
\begin{equation}
\begin{aligned}
&  \nabla f \left  (\boldsymbol \pi^*\right)- \sum\limits_{i = 0}^m \varphi_t^{\prime}\left(g_{{C_i}}\left(\boldsymbol \pi^*\right)\right) \nabla g_{{C_i}}\left(\boldsymbol \pi^*\right) +\eta \sum\limits_{i = 0}^m \nabla I^{\pi_*}_{C_i}\approx 0, \\
\label{optimality condition}
\end{aligned}
\end{equation}
where \(\nabla I^{\pi^*}(s, a)\) denotes the gradient of the intrinsic reward with respect to the policy parameters at \(\pi^*\).
The derivative of each logarithmic barrier extension $\varphi_\tau(g_{{C_i}}(\pi) )$ can be corresponding to the deterministic dual variable $\lambda_i ^*$:
\begin{equation}
\begin{aligned}
\lambda_i^*=\varphi_\tau^{\prime}\left(g_{{C_i}}\left(\boldsymbol \pi^*\right)\right). 
\label{ the derivative of the log-barrier extension(15)}
\end{aligned}
\end{equation}
Eq. (\ref{optimality condition}) implies that $\boldsymbol \pi^*$ approximately satisfies the optimality condition for the Lagrangian associated with the original inequality-constrained problem, as delineated in Eq. \eqref{eq:CMDP Optimization problem}, when $\lambda=$ $\lambda_i^*$.
\begin{equation}
\begin{aligned}
\nabla J\left(\boldsymbol \pi^*\right)-\sum\limits_{i = 0}^m \lambda_i^* \nabla g_{{C_i}}\left(\boldsymbol \pi^*\right)+\eta\sum\limits_{i = 0}^m \nabla I^{\pi^*} \approx 0.
\label{the optimality condition for the Lagrangian corresponding(16)}
\end{aligned}
\end{equation}
In addition, we can confirm that in Eq.(12), the implicit variable $\lambda_i^*$ corresponds to a feasible dual solution. This implies that every element of $\lambda_i^*$ is strictly greater than zero ($\lambda_i^* > 0$). Therefore, we can express the value of the dual function when $\lambda_i^* > 0$ as:
\begin{equation}
\begin{aligned}
G\left(\boldsymbol{\lambda}^*\right)=J\left(\boldsymbol \pi^*\right)- \sum\limits_{i = 0}^m \lambda_i^* g_{{C_i}}\left(\boldsymbol \pi^*\right)+\eta \sum\limits_{i = 0}^m \mathbb{E}\large[I^{\pi^*} \large],
\label{the dual function evaluated(17)}
\end{aligned}
\end{equation}
where \(\mathbb{E}\left[I^{\pi^*}(s, a)\right]\) is the expected value of the intrinsic reward under \(\pi^*\).
The duality gap linked to the primal-dual pair $\left(\boldsymbol{\pi}^*, \boldsymbol{\lambda}^*\right)$ is: 
\begin{equation}
\begin{aligned}
G\left(\boldsymbol{\lambda}^*\right)-J\left(\boldsymbol \pi^*\right)=- \sum\limits_{i = 0}^m \lambda_i^* g_{{C_i}}\left(\boldsymbol \pi^*\right)+\eta\sum\limits_{i = 0}^m \mathbb{E}[I^{\pi^*}].
\label{duality gap associate(18)}
\end{aligned}
\end{equation}
This duality gap between the optimal objective values of the primal and dual problems, we will consider three cases for $-\lambda_i^* g_{{C_i}}\left(\boldsymbol \pi^*\right)$:
\begin{itemize}
    \item -$g_{{C_i}}\left(\boldsymbol \pi^*\right) \leq-\frac{1}{\tau^2}$: if $\lambda_i^*=-\frac{1}{\tau g_{{C_i}}\left(\boldsymbol \pi^*\right)}$, we can derive that $-\lambda_i^* g_{{C_i}}\left(\boldsymbol \pi^*\right) \leq \frac{1}{\tau}$.
    \item -$\frac{1}{\tau^2} \leq g_{{C_i}}\left(\boldsymbol \pi^*\right) \leq 0$:  
    In this case, if $\lambda_i^*=\tau$, then we are able to derive $-\lambda_i^* g_{{C_i}}\left(\boldsymbol \pi^*\right) \leq-\lambda_i^*\left(-\frac{1}{\tau^2}\right) \leq \frac{1}{\tau}$.
    \item $g_{{C_i}}\left(\boldsymbol \pi^*\right) \geqslant 0$:
     Since $\tau$ is strictly positive, we can indeed simplify the inequality $-\lambda_i^* g_{{C_i}}\left(\boldsymbol \pi^*\right)=-\tau g_{{C_i}}\left(\boldsymbol \pi^*\right) \leq 0 \leq \frac{1}{\tau}$.
\end{itemize}
For the intrinsic reward term, since \(I^{\pi}\) is constructed to be non-negative and \(\eta > 0\), we have \(\eta \mathbb{E}\left[I^{\pi^*}\right] \geq 0\). Moreover, the intrinsic reward is scaled by a softmax function, restricting it to \((0, 1]\), so \(\mathbb{E}\left[I^{\pi^*}\right] \leq I^{\text{max}}\) for some constant \(I^{\text{max}}\), leading to \(\eta \mathbb{E}\left[I^{\pi^*}\right] \leq \eta I^{\text{max}}\).

Therefore, by combining the aforementioned three cases, we can obtain an upper bound of the duality gap between the primal problem and the dual problem given by
\begin{equation}
\begin{aligned}
G\left(\boldsymbol{\lambda}^*\right)-J\left(\boldsymbol \pi^*\right) \leq \frac{m}{\tau}+\eta\sum\limits_{i = 0}^m  I_{\max}.
\label{upper bound of the duality gap}
\end{aligned}
\end{equation}
where $\boldsymbol{\pi}^*$ is the optimal policy, and \(I_{\text{max}}\) is the upper bound of the expected intrinsic reward. The optimal implicit Lagrangian dual variables satisfy 

\begin{equation}
\begin{aligned}
& \lambda_i^*=\left\{\begin{array}{cc}
-\frac{1}{\tau g_{{C_i}}\left(\boldsymbol \pi^*\right)} & g_{{C_i}}\left(\boldsymbol \pi^*\right) \leq-\frac{1}{\tau^2}, \\
\tau & \text { otherwise. }
\end{array}\right. \\
\end{aligned}
\end{equation}

\section{Proof of Theorem 3}\label{Proof of Theorem PCPO Update Performance} 

\textit{Proof}. The primary objective establishes the discrepancy arising from two successive iterations of the barrier function
\begin{equation}\label{eq:PCPO iteration difference}
\begin{aligned}
    &G({\pi _{k + 1}}) - G({\pi _k})
\\=&f({\pi _{k + 1}}) - f({\pi _k})- (\sum\limits_{i = 1}^m \!{{\varphi _\tau }({g_{{C_i}}}({\pi _{k + 1}}))}- \sum\limits_{i = 1}^m \!\!{{\varphi _\tau }({g_{{C_i}}}({\pi _k}))} )
 +\eta\sum\limits_{i = 1}^m[I^{\pi_{k+1}}_{C_i}-I^{\pi_{k}}_{C_i}].
\end{aligned}
\end{equation}

Given an extended log-barrier function:
\begin{equation}\label{eq:continous barrier function 2}
    {\varphi _\tau(g_{{C_i}}(\pi)) \!= \!\left\{
    \begin{array}{rcl}
    -\frac{1}{\tau} \log (-g_{{C_i}}(\pi) ) & &  \!\!\! \!\!\!g_{{C_i}}(\pi)  \!\leqslant\!-\frac{1}{\tau^2}, \\
    \!\! \tau g_{{C_i}}(\pi) \!-\!\frac{1}{\tau} \log \left(\frac{1}{\tau^2}\right)\!+\!\frac{1}{\tau}  & &  \!\!\!\text { otherwise. }
\end{array}\right.}
\end{equation}

For the intrinsic reward term \(\eta \left[ I^{\pi_{k+1}}_{C_i}- I^{\pi_{k}}_{C_i}\right]\): By definition, \(I^{\pi}\) is a non-negative constraint-aware reward activated near the constraint boundary, with \(0 < I^{\pi}_{C_i} \leq I^{\max}_{C_i}\) (bounded by the softmax and sigmoid functions, where \(I^{\text{max}}_{C_i}\) is a constant upper bound). Thus, \(I^{\pi_{k}}_{C_i}- I^{\pi_{k+1}}_{C_i} \leq I^{\max}_{C_i}\), so:
\begin{equation}\label{Ik+1-Ik}
    \eta(I^{\pi_{k+1}}_{C_i} - I^{\pi_{k}}_{C_i})\geq -\eta I^{\max}_{C_i}
\end{equation}

Case 1: Considering $ {g_{{C_i}}(\pi_k) \leqslant  -\frac{1}{\tau^2}}$.

Based on  Eq. (\ref{eq:continous barrier function 2}), we can obtain
\begin{equation} \label{eq:barrier function difference}
\begin{array}{l} 
 \varphi \left(g_{{C_i}}\left(\pi_k\right)\right)-\varphi \left(g_{{C_i}}\left(\pi_{k+1}\right)\right)\\
 =\frac{1}{\tau} \log \left(-g_{{C_i}}\left(\pi_{k+1}\right)\right)-\frac{1}{\tau} \log \left(-g_{{C_i}}\left(\pi_k\right)\right)\\
 =-\frac{1}{\tau} \log \left(\frac{J_{C_i}\left(\pi_k\right)+\frac{1}{1-\gamma}  \mathop  \mathbb{E} \left[A_{C_i}^{\pi_k}\right]-{d_i}}{J_{C_i}\left(\pi_{k+1}\right)+\frac{1}{1-\gamma} \mathop  \mathbb{E} \left[A_{C_i}^{\pi_{k+1}}\right]-{d_i}}\right)\\ 
 =\!{-\frac{1}{\tau} \log \!\left(\!1\!-\!\frac{J_{C_i}\!\left(\pi_{k+1}\!\!\right)-J_{C_i}\left(\pi_k\right)+\frac{1}{1-\gamma}\left[ \mathop  \mathbb{E} \left[A_{C_i}^{\pi_{k+1}}\right]- \mathop  \mathbb{E} \left[A_{C_i}^{\pi_k}\right]\right]}{J_{C_i}\left(\pi_{k+1}\right)+\frac{1}{1-\gamma} \mathop  \mathbb{E} \left[A_{C_i}^{\pi_{k+1}}\right]-{d_i}}\!\!\!\right)}.
\end{array}
\end{equation}

\begin{equation}\label{eq:log summation equation}
\begin{array}{l}
\sum\limits_{i = 1}^m {\varphi \left( {{g_{{C_i}}}\left( {{\pi _k}} \right)} \right)}  - \sum\limits_{i = 1}^m {\varphi \left( {{g_{{C_i}}}\left( {{\pi _{k + 1}}} \right)} \right)} 
 = \sum\limits_{i = 1}^m {[\varphi \left( {{g_{{C_i}}}\left( {{\pi _k}} \right)} \right) - \varphi \left( {{g_{{C_i}}}\left( {{\pi _{k + 1}}} \right)} \right)]}. 
\end{array}
\end{equation}
Thus,
\begin{equation}\label{eq:Barrier functions are differentiated and summed}
\begin{array}{l}
\sum\limits_{i = 1}^m {\varphi \left( {{g_{{C_i}}}\left( {{\pi _k}} \right)} \right)}  - \sum\limits_{i = 1}^m {\varphi \left( {{g_{{C_i}}}\left( {{\pi _{k + 1}}} \right)} \right)} \\
 = \!\! \!- \frac{1}{\tau }\!\sum\limits_{i = 1}^m \!{\log \!\left( \!{1 \!-\! \frac{{{J_{C_i}}\!\!\left( {{\pi _{\!k + 1\!\!}}} \right) - \!{J_{C_i}}\!\left( {{\pi _k}} \right) + \frac{1}{{1 - \gamma }}\left[ {\mathbb{E}\left[ {A_{C_i}^{{\pi _{k + 1}}}} \right] - \mathbb{E}\left[ {A_{C_i}^{{\pi _k}}} \right]} \right]}}{{{J_{C_i}}\left( {{\pi _{k + 1}}} \right) + \frac{1}{{1 - \gamma }}\mathbb{E}\left[ {A_{C_i}^{{\pi _{k + 1}}}} \right] - {d_i}}}} \!\right)}. 
\end{array}
\end{equation}

Next, simplify Eq. (\ref{eq:Barrier functions are differentiated and summed}). Given that $J_{C_i}\left(\pi_k\right) \geqslant 0,\forall k$, it follows that $J_{C_i}\left(\pi_{k+1}\right) \geqslant  0$. The Proposition 2 \cite{achiam2017cpo} establishes an upper bound on the constraint violation for the CPO update. When considering the strategy constraint, we modify the KL divergence constraint to \(\mathbb{E}_{s \sim d_{\pi_k}}[D_{KL}(\pi\parallel \pi_k)] \leq \delta\). Following the same reasoning as in the original proof and utilizing Corollary 2 and Corollary 3 from \cite{achiam2017cpo}, we obtain the adjusted upper bound on constraint violation as 
\begin{equation}\label{eq:J_k upper bound}
     J_{C_i}(\pi_{k+1}) \leq d_i + \frac{\sqrt{2\delta}\gamma\epsilon_{C_i}^{\pi_{k+1}}}{(1-\gamma)^2},
\end{equation}
where the point $\pi_k$ is a feasible solution for \eqref{eq:Simplified objective function} with an objective value of 0 from the PCPO proposition 1 \cite{achiam2017cpo}, and $\quad \epsilon^{\pi_k}={\max _s} {\left|   {{\mathbb{E}_{a\sim{\pi _{k}}}}\left[ {{A^{{\pi _k}}}} \right]} \right|} = 0$, $\quad \epsilon_{C_i}^{\pi_k}={\max _s} {\left|   {{\mathbb{E}_{a\sim{\pi _{k}}}}\left[ {{A_{C_i}^{{\pi _k}}}} \right]} \right|}\ne 0$, therefore,    
\begin{equation}\label{eq:constraint difference}
{J_{C_i}}\left( {{\pi _{k + 1}}} \right) - {J_{C_i}}\left( {{\pi _k}} \right) \geqslant -{d_i}-\frac{\sqrt{2 \delta} \gamma  \epsilon_{C_i}^{\pi_k}}{(1-\gamma)^2}.
\end{equation}
Due to
\begin{equation}\label{eq5}
    {J\left(\pi_{k+1}\right)-J\left(\pi_k\right)=\frac{1}{1-\gamma}\left[ \mathop  \mathbb{E} \left[A^{\pi_{k+1}}\right]- \mathop  \mathbb{E} \left[A^{\pi_k}\right]\right]}.
\end{equation}
According to Proposition 1, Corollary 1, and Corollary 3 of PCPO \cite{achiam2017cpo}, we have the following bound holds:
\begin{equation}\label{eq:PCPO proposition 1}
\begin{array}{l}
    {J\left( {{\pi _{k + 1}}} \right) - J\left( {{\pi _k}} \right) }
    {\geqslant \frac{{ - \sqrt {2\delta} \gamma {\epsilon^{{\pi _{k + 1}}}}}}{{{(1 - \gamma )}^2}},{\epsilon^{{\pi _{k + 1}}}} = {\max _s} {\left|   {{\mathbb{E}_{a\sim{\pi _{k + 1}}}}\left[ {{A^{{\pi _k}}}} \right]} \right|}}.
\end{array}
\end{equation}
Similarly, we can obtain the following equation:

\begin{equation}\label{eq:Constrained advantage function difference}
    {\frac{1}{1-\gamma}\left[ \mathop  \mathbb{E} \left[A_{C_i}^{\pi_{k+1}}\right]- \mathop  \mathbb{E} \left[A_{C_i}^{\pi_k}\right]\right] \geqslant \frac{-\sqrt{2 \delta} \gamma \epsilon_{C_i}^{\pi_{k+1}}}{(1-\gamma)^2} }.
\end{equation}
By applying Holder’s inequality; for any $p, q \in [1, \infty]$ such that $1/p+1/q = 1$, we have:
\begin{equation}\label{eq:advantage_function_upper_bound}
\begin{split}
    &\frac{1}{1-\gamma} \mathop{\mathbb{E}} \left[A_{C_i}^{\pi_{k+1}}\right] \leqslant \left\|d_{\pi_{k+1}} - d_{\pi_k}\right\|_p \left\|\mathbb{E}_{{a \sim \pi_{k+1}} \atop {s^{\prime} \sim \pi_{k+1}}}\left[A_{C_i}^{\pi_{k+1}}\right]\right\|_q   = \frac{\sqrt{2 \delta} \gamma \epsilon_{C_i}^{\pi_{k+1}}}{(1-\gamma)^2}.
\end{split}
\end{equation}

Substituting Eq.(\ref{eq:log summation equation})$\sim$Eq.(\ref{eq:advantage_function_upper_bound}) into Eq.(\ref{eq:barrier function difference}),from this, we can infer:

\begin{equation}\label{eq11}
\begin{aligned}
& \varphi\left(g_{{C_i}}\left(\pi_k\right)\right)-\varphi\left(g_{{C_i}}\left(\pi_{k+1}\right)\right)\\
& \geqslant -\frac{1}{\tau} \log \left(1+\frac{{d_i}(1-\gamma)^2+\sqrt{2 \delta} \gamma (\epsilon_{C_i}^{\pi_{k}}+\epsilon_{C_i}^{\pi_{k+1}})}{2 \sqrt{2 \delta} \gamma \epsilon_{C_i}^{\pi_{k+1}}}\right)\\
&\approx -\frac{1}{\tau} \log \left(2+\frac{{d_i}(1-\gamma)^2}{2 \sqrt{2 \delta} \gamma \epsilon_{C_i}^{\pi_{k+1}}}\right).
\end{aligned}
\end{equation}
For case 1, the difference between the two iterative updates is
\begin{equation}\label{eq: the difference between the two iterative updates}
\begin{aligned}
&f({\pi _{k + 1}})\! \!-\!\! f({\pi _k})\! - \!\!(\sum\limits_{i = 1}^m {{\varphi _\tau }({g_{{C_i}}}({\pi _{k + 1}}))}  \!- \!\sum\limits_{i = 1}^m {{\varphi _\tau }({g_{{C_i}}}({\pi _k}))} )\\
&\geqslant \frac{{ - \sqrt {2\delta} \gamma \epsilon^{{\pi _{k + 1}}}}}{{{{(1 - \gamma )}^2}}} - \frac{1}{\tau }\sum\limits_{i = 1}^m {\log \left( {2 + \frac{{{d_i}{{(1 - \gamma )}^2}}}{{2\sqrt {2\delta } \gamma \epsilon_{C_i}^{{\pi _{k + 1}}}}}} \right)}. 
\end{aligned}
\end{equation}

Case 2: Considering ${ g_{{C_i}}(\pi_k) >-\frac{1}{\tau^2}}$:
\begin{equation}\label{eq:Barrier functions are differentiated and summed in case 2}
\begin{array}{l}
\sum\limits_{i = 1}^m {\varphi \left( {{g_{{C_i}}}\left( {{\pi _k}} \right)} \right)} \!\! - \sum\limits_{i = 1}^m {\varphi \left( {{g_{{C_i}}}\left( {{\pi _{k + 1}}} \right)} \right)} \\
=\tau\left(g_{{C_i}}\left(\pi_k\right)-g_{{C_i}}\left(\pi_{k+1}\right)\right)\\
 = \!{\tau }\!\sum\limits_{i = 1}^m \!\!\left[J_{C_i}\!\!\left(\pi_k\right)\!-\!\!J_{C_i}\!\!\left(\pi_{k+1}\right)\!+\!\frac{1}{1-\gamma}\!\left[ \mathop  \mathbb{E} \!\left[A_{C_i}^{\pi_k}\right] \!- \!\mathop  \mathbb{E} \!\left[A_{C_i}^{\pi_{k+1}}\right]\right]\!\right].
\end{array}
\end{equation}
It can be obtained from Eq.\eqref{eq:constraint difference}
\begin{equation}\label{Jck-Jck+1}
{J_{C_i}}\left( {{\pi _{k}}} \right) - {J_{C_i}}\left( {{\pi _{k+1}}} \right) \le {d_i}+\frac{\sqrt{2 \delta/\alpha} \gamma  \epsilon_{C_i}^{\pi_k}}{(1-\gamma)^2} .
\end{equation}
Equation \eqref{eq:Constrained advantage function difference} can be rewritten as

\begin{equation}\label{eq:Constrained advantage function difference in case 2}
    {\frac{1}{1-\gamma}\left[ \mathop  \mathbb{E} \left[A_{C_i}^{\pi_{k}}\right]- \mathop  \mathbb{E} \left[A_{C_i}^{\pi_{k+1}}\right]\right] \le \frac{\sqrt{2 \delta} \gamma \epsilon_{C_i}^{\pi_{k+1}}}{(1-\gamma)^2} }.
\end{equation}

By Eq. (\ref{eq:Barrier functions are differentiated and summed in case 2}) $\sim$ Eq. (\ref{eq:Constrained advantage function difference in case 2}), $\text{we can infer:}$
\begin{equation}\label{eq:Barrier_functions_case_2}
\begin{aligned}
    & \sum_{i = 1}^m \varphi \left( g_{C_i}(\pi_k) \right) - \sum_{i = 1}^m \varphi \left( g_{C_i}(\pi_{k + 1}) \right) \leq \tau m d_i + \frac{\sqrt{2\delta} \tau \gamma \sum_{i = 1}^m (\epsilon_{C_i}^{\pi_k} + \epsilon_{C_i}^{\pi_{k + 1}})}{(1 - \gamma)^2}.
\end{aligned}
\end{equation}

For case 2, based on Eq. (\ref{eq:PCPO proposition 1}) and Eq. (\ref{eq:Barrier functions are differentiated and summed in case 2}), we can get the difference between the two iterative updates is
\begin{equation}\label{eq: the difference between the two iterative updates in case 2}
\begin{aligned}
&f({\pi _{k + 1}})\! - \!f({\pi _k})\! - \!(\sum\limits_{i = 1}^m {{\varphi _\tau }({g_{{C_i}}}({\pi _{k + 1}}))}\!  - \!\sum\limits_{i = 1}^m {{\varphi _\tau }({g_{{C_i}}}({\pi _k}))} )\\
&\geqslant \frac{{ - \sqrt {2\delta} \gamma \epsilon^{{\pi _{k + 1}}}}}{{{{(1 - \gamma )}^2}}} - \tau m{d_i} - \frac{{\sqrt {2\delta} \tau \gamma \sum\limits_{i = 1}^m (\epsilon_{{C_i}}^{{\pi _k}} + \epsilon_{{C_i}}^{{\pi _{k + 1}}})}}{{{{(1 - \gamma )}^2}}}\\
&= - \tau m{d_i} - \frac{{\sqrt {2\delta} \gamma \sum\limits_{i = 1}^m ( \epsilon^{{\pi _{k + 1}}}+\tau \epsilon_{{C_i}}^{{\pi _k}}+\tau \epsilon_{{C_i}}^{{\pi _{k + 1}}})}}{{{{(1 - \gamma )}^2}}}.
\end{aligned}
\end{equation}
From Eq.\eqref{eq: the difference between the two iterative updates} in case 1 and Eq.\eqref{eq: the difference between the two iterative updates in case 2} in case 2, we can derive the following:
\begin{equation}\label{eq20}
\begin{aligned}
&f(\!{\pi _{k + 1}}\!) \!- \!f({\pi _k}) \!- \!(\sum\limits_{i = 1}^m {{\varphi _\tau }({g_{{C_i}}}({\pi _{k + 1}}))}  - \sum\limits_{i = 1}^m {{\varphi _\tau }({g_{{C_i}}}({\pi _k}))} )\\
&\geqslant \!\!\left\{ \!\!\!{\begin{array}{*{20}{c}}
{\frac{{ - \sqrt {2\delta} \gamma \epsilon_{}^{{\pi _{k + 1}}}}}{{{{(1 - \gamma )}^2}}} \!-\! \frac{1}{\tau }\!\!\sum\limits_{i = 1}^m \!{\log \!\left( {2 \!+\! \frac{{{d_i}{{(1 - \gamma )}^2}}}{{2\sqrt {2\delta} \gamma \epsilon_{{C_i}}^{{\pi _{k + 1}}}}}} \!\right)} ,}& \!\!\!\!{g_{{C_i}}}(\pi_k) \!\!\leqslant\!\!-\frac{1}{\tau^2},\\
{ - \tau m{d_i} \!-\! \frac{{\sqrt {2\delta} \gamma \!\sum\limits_{i = 1}^m {(\epsilon^{{\pi _{k + 1}}}+\tau \epsilon_{{C_i}}^{{\pi _k}}\! + \!\tau \epsilon_{{C_i}}^{{\pi _{k + 1}}})} }}{{{{(1 - \gamma )}^2}}},}&{{\rm{ otherwise. }}}
\end{array}} \right.
\end{aligned}
\end{equation}
where$\quad \epsilon_{C_i}^{\pi_k}={\max _s} {\left|   {{\mathbb{E}_{a\sim{\pi _{k}}}}\left[ {{A_{C_i}^{{\pi _k}}}} \right]} \right|}$ and $\epsilon_{{C_i}}^{{\pi _{k + 1}}} = {\max _s}|{\mathbb{E}_{a\sim \pi_{k + 1}}}[A_{{C_i}}^{{\pi _{k + 1}}}]|$. 
With Eq.\eqref{Ik+1-Ik}, Eq.(\ref{eq20}) can be further simplified  as
\begin{equation}\label{eq:21}
\small\begin{aligned}
&G({\pi _{k + 1}}) - G({\pi _k}) \geqslant\\
& \!\left\{ {\begin{array}{*{20}{c}}
{\!\!\!\!{\eta }({\pi _{k + 1}})\!- \!\frac{1}{\tau }\sum\limits_{i = 1}^m {\log \left( {2\! - \!\frac{{{d_i}}}{{2{\eta _{{C_i}}}({\pi _{k + 1}})}}} \right)-\eta\sum\limits_{i = 1}^m I^{\max}_{C_i}} },{g_{\!{C_i}\!}}(\pi_k) \!\leqslant\!-\!\frac{1}{\tau^2},\\
{ - m{d_i}\! \!\!+ \!\!{\eta}({\pi _{k + 1}})\!\!+\!\!\sum\limits_{i = 1}^m {( {\eta _{{C_i}}}({\pi _k})\! + \!{\eta _{{C_i}}}({\pi _{k + 1}}))-\eta\sum\limits_{i = 1}^m I^{\max}_{C_i}}},{{\rm{ otherwise. }}}
\end{array}} \right.
\end{aligned}
\end{equation}
where ${\eta}({\pi _{k + 1}}) = \frac{{ - \sqrt {2\delta} \gamma \epsilon^{{\pi _{k + 1}}}}}{{{{(1 - \gamma )}^2}}}$, ${\eta _{{C_i}}}({\pi _k}) = \frac{{ - \sqrt {2\delta} \gamma \epsilon_{{C_i}}^{{\pi _{k }}}}}{{{{(1 - \gamma )}^2}}}$, and ${\eta _{{C_i}}}({\pi _{k + 1}}) = \frac{{ - \sqrt {2\delta} \gamma \epsilon_{{C_i}}^{{\pi _{k + 1}}}}}{{{{(1 - \gamma )}^2}}}$.


\section{Proof of Proposition 4}
\textit{Proof}. 
To establish \(V_P(T) \leq V_L(T) - \Delta(T)\) where \(\Delta(T) > 0\) and \(\Delta(T) \to \infty\) as \(T \to \infty\), we analyze the fundamental differences between PCPO and Lagrangian-based methods in managing constraint violations, supplemented with rigorous mathematical derivations. For any iteration $t$, define the constraint violation as 
\begin{equation}
    v_t = \max(0, J_C(\pi_t) - d),
\end{equation}
where \(J_C(\pi_t)\) is the cumulative cost of policy \(\pi_t\) and $d$ is the cost threshold. The cumulative violations for PCPO are 
\begin{equation}
    V_P(T) = \sum_{t=1}^T v_t^P,
\end{equation}
where \(v_t^P = \max(0, J_C(\pi_t^P) - d)\).
And for Lagrangian methods, the cumulative violations are \(V_L(T) = \sum_{t=1}^T v_t^L\), where \(v_t^L = \max(0, J_C(\pi_t^L) - d)\).


PCPO employs two core components embedded in its objective function \(G(\pi_\theta) = f(\pi_\theta) - \sum_{i=1}^m \varphi_\tau(g_{C_i}(\pi_\theta)) + \eta \sum_{i=1}^m I_{C_i}^{\pi_\theta}\). The extended log-barrier function \(\varphi_\tau(g_{C_i}(\pi_\theta))\) generates a strictly positive gradient as \(g_{C_i}(\pi_\theta) \to 0^-\), creating a "repulsive force" to prevent violations, with its derivative ensuring proactive correction. The constraint-aware intrinsic reward \(I_{C_i}^{\pi_\theta}\) activates only near boundaries, incentivizing exploration within feasible regions. In contrast, Lagrangian methods use penalties that are zero when constraints are satisfied and only activate post-violation, leading to hysteresis in multiplier updates, oscillations, and overshoots. In contrast, Lagrangian methods lack boundary-awareness. As shown in Figure \ref{fig:Lagrange}, the penalty \(B(g(x)) = 0\) when \(g(x) \leq 0\) (constraints satisfied), and only reacts post-violation (\(g(x) > 0\)). This hysteresis in Lagrange multiplier updates causes oscillations and overshoots, leading to repeated violations. 

We first calculate the lower bounds of the improvement values for the PCPO and the Lagrangian-based safe RL methods and then compare these two bounds. The lower bound of the improvement values for adjacent iterations of PCPO is given in Theorem 2. Next, we derive the lower bound of the improvement values for the Lagrangian-based safe RL method. The Lagrange multiplier term is defined as follows:
\begin{equation}\label{Lagrange multipliers}
{\varphi  _\tau }({g_{{C_i}}}(\pi_k )) = \lambda {g_{{C_i}}}(\pi_k ).
\end{equation}
The difference arises from two consecutive iterations of the Lagrangian function:
\begin{equation}\label{eq:Lagrangian iteration difference}
\begin{array}{l}
G({\pi _{k + 1}}) - G({\pi _k})\\
 = \!f({\pi _{k + 1}})\!\! - \!\!f({\pi _k})\!\! - \!\!(\sum\limits_{i = 1}^m \!{{\varphi _\tau }({g_{{C_i}}}({\pi _{k + 1}}))}\!\!  - \!\!\sum\limits_{i = 1}^m \!\!{{\varphi _\tau }({g_{{C_i}}}({\pi _k}))} )\\
 = \!f({\pi _{k + 1}})\!\! - \!\!f({\pi _k})\!\!  - \lambda_i \sum\limits_{i = 1}^m  ({g_{{C_i}}}({\pi _{k + 1}}) - {g_{{C_i}}}({\pi _k})).
\end{array}
\end{equation}
By applying Holder’s inequality; for any $p, q \in [1, \infty]$ such that $1/p+1/q = 1$, we have:
\begin{equation}\label{eq:advantage_function_upper_bound}
\begin{split}
    & \frac{1}{1-\gamma} \mathop{\mathbb{E}} \left[A_{C_i}^{\pi_{k+1}}\right] \geqslant -\left\|d_{\pi_{k+1}} - d_{\pi_k}\right\|_p \left\|\mathbb{E}_{{a \sim \pi_{k+1}} \atop {s^{\prime} \sim \pi_{k+1}}}\left[A_{C_i}^{\pi_{k+1}}\right]\right\|_q= -\frac{\sqrt{2 \delta} \gamma \epsilon_{C_i}^{\pi_{k+1}}}{(1-\gamma)^2}.
\end{split}
\end{equation}
Based on Eq.(\ref{eq:Constrained advantage function difference}) and Eq.(\ref{eq:advantage_function_upper_bound}), we obtain 
\begin{equation}\label{eq:advantage_function_bound}
 - \frac{{\sqrt {2\delta } \gamma \epsilon_{{C_i}}^{{\pi _k}}}}{{{{(1 - \gamma )}^2}}} \le \frac{1}{{1 - \gamma }}\mathop{\mathbb{E}}\left[ {A_{{C_i}}^{{\pi _k}}} \right] \le \frac{{\sqrt {2\delta } \gamma \epsilon_{{C_i}}^{{\pi _k}}}}{{{{(1 - \gamma )}^2}}}
\end{equation}
From Eq.(\ref{eq:J_k upper bound}), Eq.(\ref{eq:advantage_function_upper_bound}), and Eq.(\ref{eq:advantage_function_bound}), we analyze the change in the constraint function \(g_{C_i}\) between iterations:
\begin{equation}\label{eq:Lagrangian iteration difference}
\begin{array}{l}
{g_{{C_i}}}({\pi _{k + 1}}) - {g_{{C_i}}}({\pi _k})\\
=J_{C_i}\left(\pi_{k+1}\right)-J_{C_i}\left(\pi_k\right)+\frac{1}{1-\gamma}\left[ \mathop  \mathbb{E} \left[A_{C_i}^{\pi_{k+1}}\right]- \mathop  \mathbb{E} \left[A_{C_i}^{\pi_k}\right]\right]\\
 \le {d_i} + \frac{{\sqrt {2\delta } \gamma (2_{\epsilon_{C_i}}^{{\pi _{k + 1}}} + _{\epsilon_{C_i}}^{{\pi _k}})}}{{{{(1 - \gamma )}^2}}}\\
 = {d_i} - {\eta _{{C_i}}}({\pi _k}) - 2{\eta _{{C_i}}}({\pi _{k + 1}}).
\end{array}
\end{equation}
Here, \(\eta_{C_{i}}(\pi) = \frac{-\sqrt{2 \delta} \gamma \epsilon_{C_{i}}^{\pi}}{(1-\gamma)^{2}}\) (negative due to \(\epsilon_{C_{i}}^{\pi} > 0\)), clarifying the substitution from the cost advantage bounds to the \(\eta\)-term.
Thus, substituting this result into the Lagrangian update difference, we get:
\begin{equation}\label{eq:G difference}
\begin{array}{l}
G({\pi _{k + 1}}) - G({\pi _k})\\
\geqslant {\eta}({\pi _{k + 1}}) - \lambda_i m{d_i} + \lambda_i \sum\limits_{i = 1}^m {( {\eta _{{C_i}}}({\pi _k})\! + \!{2\eta _{{C_i}}}({\pi _{k + 1}}))}.
\end{array}
\end{equation}
From Eq.(12), the optimal dual variable $\lambda _i^* \le \tau $. As $\lambda_i$ increases, the lower bound continuously decreases because the term \(-\lambda_i m d_i\) dominates. Therefore, due to the term containing \(\tau\) being negative, the minimum lower bound is achieved when $\lambda _i^* = \tau$.
\begin{equation}\label{eq:max lower bound}
\begin{array}{l}
\min_{\lambda_i} \left(\!\!\! \eta({\pi_{k + 1}}) \!\!- \!\!\lambda_i m d_i \!+\! \lambda_i \sum\limits_{i = 1}^m \left( \eta_{C_i}({\pi_k}) \!+\! 2 \eta_{C_i}({\pi_{k + 1}}) \right) \!\! \right) \\
= \eta({\pi_{k + 1}}) - \tau m d_i + \tau \sum\limits_{i = 1}^m \left( \eta_{C_i}({\pi_k}) + 2 \eta_{C_i}({\pi_{k + 1}}) \right).
\end{array}
\end{equation}
Then, we define the lower bound of update performance as:
    \begin{equation}
        \begin{aligned}
       & L_\text{PCPO}= {\begin{cases} 
  \eta(\pi_{k+1}) - \frac{1}{\tau} \sum_{i=1}^m \log\left(2 - \frac{d_i}{2\eta_{C_i}(\pi_{k+1})}\right) - \eta \sum_{i=1}^m I_{C_i}^{\text{max}},\\
  -m d_i + \eta(\pi_{k+1}) + \sum_{i=1}^m \left(\eta_{C_i}(\pi_k) + \eta_{C_i}(\pi_{k+1})\right) - \eta \sum_{i=1}^m I_{C_i}^{\text{max}}.
  \end{cases}}
    \end{aligned} 
    \end{equation}
Moreover, for Lagrangian-based safe RL methods, we define the update performance lower bound \(L_{Lagrange}\) as
\begin{equation}
    \begin{aligned}
       & L_\text{Lagrange} = {\small\eta(\pi_{k+1}) - \tau m d_i + \tau \sum_{i=1}^m \left(\eta_{C_i}(\pi_k) + 2\eta_{C_i}(\pi_{k+1})\right)}.
    \end{aligned}
\end{equation}
Thus, we can obtain
\begin{equation}
    L_{\text{Lagrange}} < L_{\text{PCPO}},
\end{equation}
Define the difference between the lower bounds of the two in each update as $\delta_k$, a positive number:
\begin{equation}
\delta_k \;=\; L_\text{PCPO}-L_\text{Lagrange} \;>\;0.
\end{equation}
This gap arises inherently from PCPO's proactive mechanisms—its preemptive barrier penalties and constraint-aware intrinsic rewards enable more efficient balancing of reward maximization and constraint satisfaction compared to Lagrangian methods, which rely on reactive penalty adjustments.
Since \(L_{\text{PCPO}}\) and \(L_{\text{Lagrange}}\) measure the minimum improvement in the objective function (trading off reward gains and constraint costs), a larger \(L_{\text{PCPO}}\) implies that PCPO achieves better control over constraint violations for equivalent reward improvements. Consequently, for any iteration k, the cumulative constraint cost under PCPO's updated policy \(\pi^{P}_{k+1}\) must be bounded above by that of the Lagrangian-updated policy \(\pi^{L}_{k+1}\) minus the gap \(\delta_k\), leading to the relation:
\begin{equation}
    J_C(\pi_{k+1}^{P}) \;\le\;
J_C(\pi_{k+1}^{L}) \;-\;\delta_k,
\end{equation}
that is,
\begin{equation}
\begin{aligned}
       v_{k+1}^{P}
= &J_C(\pi_{k+1}^{P})-d
\;\\ \le &
J_C(\pi_{k+1}^{L})-d
\;-\delta_k
= v_{k+1}^{L}-\delta_k
\end{aligned}
 \end{equation}

\begin{figure}[tbp] 
    \vskip 0.2in
    \begin{center}    
    \centering{\includegraphics[width=1\linewidth]{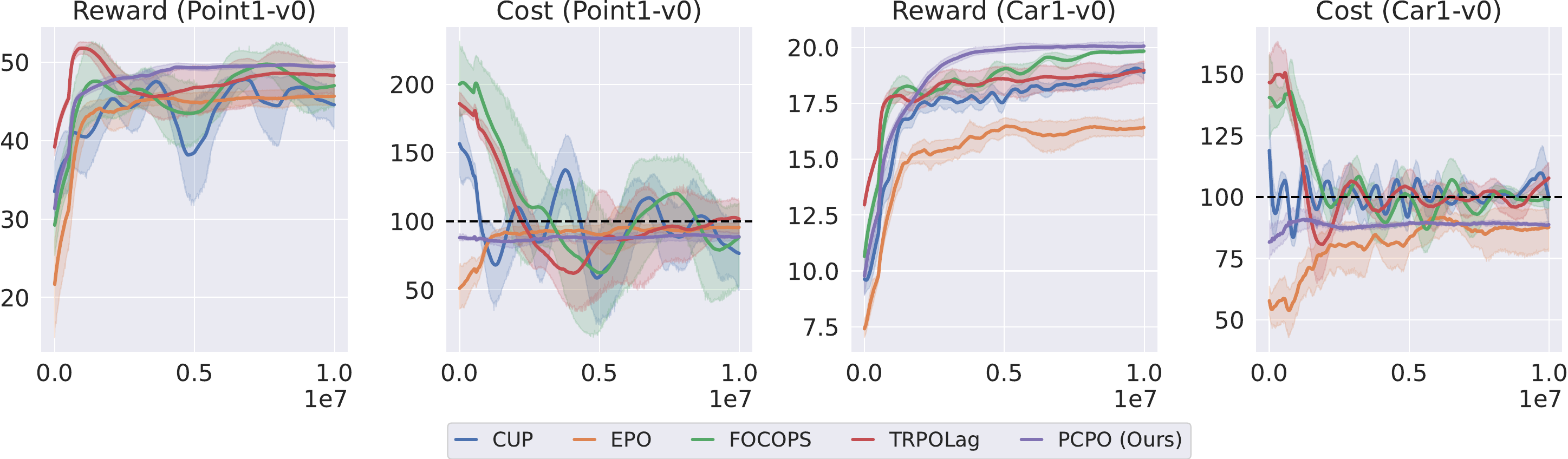}}
    \caption{The average performance for CUP, EPO, FOCOPS, TRPOLag, and PCPO on circle tasks (about 1000 bootstrap samples over 6 random seeds). The x-axis indicates the total number of samples, while the y-axis shows the mean total reward/cost return from the most recent 100 episodes. The shaded areas indicate the 95$\%$ confidence interval.}
    \label{figcircle}
    \end{center}
\end{figure}

This violation gap arises from mechanistic differences:  When \(g_{C_i}(\pi_t) \leq -1/\tau^2\), PCPO’s barrier function generates a positive gradient \(\nabla \varphi_\tau > 0\) (Eq. 6) to push the policy \(\pi_t\) away from violation, while Lagrangian gradients vanish; When \(g_{C_i}(\pi_t) > 0\) (violated), PCPO's barrier and intrinsic reward accelerate correction, whereas Lagrangian multipliers adjust sluggishly.
Thus, for \(\delta_t > 0\), there is 
\begin{equation}
    \mathbb{E}[v_t^P] \leq \mathbb{E}[v_t^L] - \delta_t,
\end{equation}
with \(\delta_t\) scaling with the duality gap reduction (Theorem 1) and intrinsic reward efficacy. 
Summing both sides from $k=0$ to $T-1$, we can obtain  
\begin{equation}
\begin{aligned}
     V_P(T) &= \sum_{k=1}^T \mathbb{E}[v_t^P]\leq \sum_{k=1}^T \left( \mathbb{E}[v_t^L] - \delta_t \right) = V_L(T) - \Delta(T),
\end{aligned}
\end{equation}
where \(\Delta(T) = \sum_{k=1}^T \delta_t\). 
Since \(\delta_t > 0\) for all $t$ (due to mechanistic differences), \(\Delta(T) > 0\). The term \(\Delta(T)\) correlates positively with \(T\): Lagrangian oscillations perpetuate violations, causing \(\delta_t\) to persist. As \(T \to \infty\), PCPO converges faster to feasible policies (Theorem 2), reducing \(v_t^P \to 0\), and Lagrangian methods exhibit non-vanishing average violation 
\begin{equation}
    \liminf_{T \to \infty} \frac{1}{T} V_L(T) > 0,
\end{equation}
making \(\Delta(T)\) unbounded.   Thus,  \(\Delta(T) \to \infty\) as \(T \to \infty\).

\section{Implement Details for Experiments}\label{Implement Details for Experiments}
\subsection{Experiment Environments}


\begin{table}[t] 
\centering
\begin{minipage}{\textwidth} 
    \caption{Hyper-parameters for robots.}\label{tab:hyper}
    \vskip 0.15in
    \begin{center}
    \begin{small} 
    \begin{sc}
    \label{tab:my_label}
    \setlength{\tabcolsep}{1mm}{
    \begin{tabular}{ccccccc}
        \toprule
            Hyperparameter & CUP & EPO & FOCOPS & TRPOLag & PCPO \\
            \hline
            No. of hidden layers & 2 & 2 & 2 & 2 & 2\\
            No. of hidden nodes  & 64 & 64 & 64 & 64 & 64 \\
            Activation & tanh & tanh & tanh & tanh & tanh \\
            Discount for reward $\gamma$ & 0.99 & 0.99 & 0.99 & 0.99 & 0.99 \\
            Discount for reward $\gamma_C$ & 0.99 & 0.99 & 0.99 & 0.99 & 0.99 \\
            Batch size & 10000 & 10000 & 10000 & 10000 & 10000 \\
            Minibatch size & 64 & 64 & 64 & 64 & 64 \\
            No. of update iters & 1 & 1 & 1 & 10 & 10 \\
            Maximum episode length & 1000 & 1000 & 1000 & 1000 & 1000 \\
            GAE parameter (reward) & 0.95 & 0.95 & 0.95 & 0.95 & 0.95 \\
            GAE parameter (cost) & 0.95 & 0.95 & 0.95 & 0.95 & 0.95 \\
            Learning rate for policy & $3\times10^{-4}$ & $3\times10^{-4}$ & $3\times10^{-4}$ & N/A & N/A \\
            Learning rate for reward value net & $3\times10^{-4}$ & $3\times10^{-4}$ & $3\times10^{-4}$ & $3\times10^{-4}$ & $3\times10^{-4}$ \\
            Learning rate for cost value net & $3\times10^{-4}$ & $3\times10^{-4}$ & $3\times10^{-4}$ & $3\times10^{-4}$ & $3\times10^{-4}$ \\
            $L2$-regularization coeff. for value net & $10^{-3}$ & $10^{-3}$ & $10^{-3}$ & $10^{-3}$ & $10^{-3}$\\
            Clipping coefficient & N/A & 0.2 & N/A & N/A & N/A \\
            Damping coeff. &  N/A & N/A & N/A & 0.1  & 0.1 \\
            Max conjugate gradient iterations & N/A & N/A & N/A & 15 & 15 \\
            Trust region bound $\delta$ & 0.02 & N/A & N/A & 0.01 & 0.01 \\
            Initial $\lambda$, $\lambda_{max}$ & 0, 2 & N/A & 0, 2 & 0, 2 & N/A \\
            Learning rate for $\lambda$ & 0.01 & N/A & 0.01 & 0.01 & N/A \\           
        \bottomrule
    \end{tabular}}
   \end{sc}
   \end{small} 
   \end{center}
\end{minipage}
\begin{minipage}{\textwidth} 
    \caption{\label{table2} The table shows the average return across 10 episodes for agents trained on robot speed limit tasks, tested with 10 new random seeds. We display the bootstrap mean and the standard 95$\%$ confidence interval, computed from 1000 bootstrap samples.}
 \vskip 0.1in
    \begin{small}
    \begin{sc}
         \hspace{-1cm}
    \setlength{\tabcolsep}{0.0000000001mm}{ 
    \begin{tabular}{cccccccc}
        \toprule
        Environment & & CUP & EPO & FOCOPS & TRPOLag & PCPO \\
        \hline
        Hopper-v1 & Return & $1287.69\pm21.11$ & $1304.36\pm27.25$ & $1133.02\pm17.61$ & $1237.65\pm25.9$ & $\bm{1444.13\pm31.5}$ \\
        (250.) & Cost & $267.8\pm4.02$ & $254.6\pm6.76$ & $249.1\pm4.42$ & $234.3\pm5.43$ & $252.5\pm5.23$\\
        \hline
        Walker2d-v1 & Return & $2582.0\pm501.51$  & $3030.99\pm333.98$ & $3172.67\pm63.75$ & $3445.85\pm449.53$ & $\bm{4011.1\pm20.47}$ \\
        (333.) & Cost & $367.9\pm76.15$ & $335.6\pm41.03$ & $314.65\pm40.31$ & $272.7\pm26.42$ & $331.7\pm3.66$\\
        \hline
        Ant-v1   & Return & $3704.47\pm284.92$ & $2860.45\pm107.68$ & $3589.99\pm230.98$ & $3811.68\pm22.74$ & $\bm{3950.55\pm35.78}$ \\
        (465.) & Cost & $451.4\pm106.61$ & $439.3\pm82.93$ & $481.43\pm64.17$ & $466.2\pm9.06$ & $465.9\pm8.69$ \\
        \hline
        HalfCheetah-v1 & Return & $3313.75\pm985.4$ & $3164.53\pm25.11$ & $3001.82\pm17.1$ & $2971.09\pm19.66$ & $\bm{3977.46\pm8.34}$ \\
        (450.) & Cost & $455.9\pm218.95$ & $444.1\pm6.14$ & $467.7\pm5.47$ & $221.9\pm26.3$ & $443.4\pm1.18$ \\
        \hline
        \bottomrule
    \end{tabular}}
    \end{sc}
    \end{small}
\end{minipage} 
\end{table}

\subsubsection{Safe Velocity}
The Safe Velocity tasks introduce velocity constraints for agents. Four models are applied sequentially to address comparative questions, specifically for the Hopper, Walker, Ant, and HalfCheetah, each increasing in dimensional state space and complexity of actions. The state of the robots comprises their generalized positions and velocities, and the control is implemented through joint torques. The cost function is formulated as follows:
\[
C_t = \mathbb{I}(V_{\text{current}} > V_{\text{threshold}})
\]

where $\mathbb{I}(\cdot)$ denotes the indicator function. Here, $V_{\text{current}}$ is the current speed of the agent, and $V_{\text{threshold}}$ represents the threshold speed, which is set to $50\%$ of the agent's maximum velocity achieved after training for 1 million steps using PPO algorithms. For more comprehensive information on the environment, please refer to the relevant documentation. 

\textbf{Environment:}https://safety-gymnasium.readthedocs.io/en/latest/environments/safe\_velocity.html
safety-gymnasium.readthedocs.io/en/latest/environments/safe\_velocity.html

\begin{figure}[ht]
    \begin{center}    
    \subfloat[PointCircle1-v0]{
        \includegraphics[width=.3\linewidth]{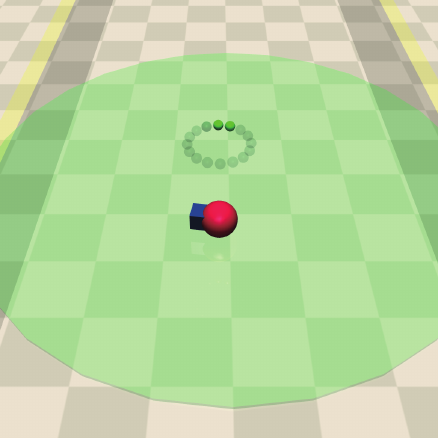}
    }
    \hspace{0.02\textwidth}  
    \subfloat[CarCircle1-v0]{
        \includegraphics[width=.3\linewidth]{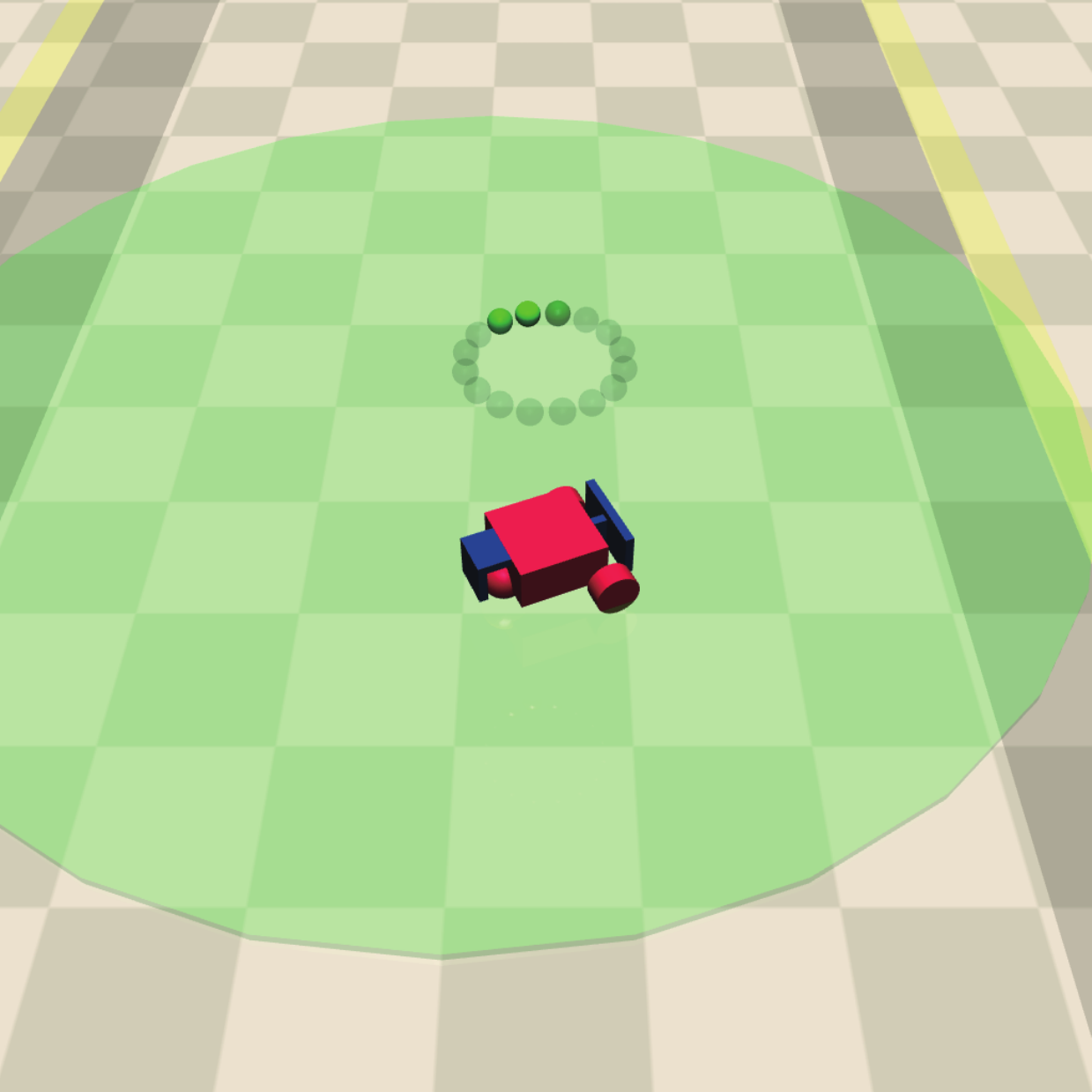}
    }
    \caption{Circle tasks}
    \label{fig:circletaskl}
    \end{center}
\end{figure}

\subsubsection{Safe Navigation}\label{safe navigation}
For the Safe Navigation tasks, we select circle tasks to evaluate our proposed PCPO. An agent's goal in these experiments is to circle around the center of the circle area while ensuring it does not cross the boundaries. We test two agent types, designated as the point and car agents, as illustrated in Figure \ref{fig:circletaskl}. The observation comprises information about the agent's current state and lidar points. The reward function $R_t$ and cost function $C_t$ are calculated as follows:
\begin{align}
    R_t &= \frac{1}{1 + |r_{agent} - r_{circle}|}\ast\frac{uy+vx}{r_{agent}}\\
    C_t &= \mathbb{I}(|x| > x_{boundary})\nonumber
\end{align}
where $x$ and $y$ denote $x-y$ axis coordinates of the agent, while $u$, $v$ is the $x-y$ axis velocity component of the agent. $r_{agent}$ represents the Euclidean distance of the agent from the origin, and $r_{circle}$ is the radius of the Circle geometry. $x_{boundary}$ is the safety margin. For more comprehensive information on the environment, please refer to:\thinspace\url{https://safety-gymnasium.readthedocs.io/en/latest/environments/safe_navigation/circle.html}
\subsubsection{Algorithmic Hyperparameters}
Table \ref{tab:hyper} summarizes the general hyperparameters used in experiments. Unless otherwise specified, we use default settings of $\tau=20$, $\alpha=0.3$, and $\beta = 1$, which consistently demonstrate strong performance across all tasks. For the intrinsic reward scaling coefficient $\alpha$, we observe that its optimal value can vary across scenarios and random seeds; therefore, we perform tuning for each setting individually.

\subsection{Adaptability Analysis Under Cost Thresholds}\label{ap8}
\begin{figure}[htbp]
\begin{minipage}{\textwidth} 
    \begin{center}    
    \centering{\includegraphics[width=1\linewidth, height=0.5\linewidth]{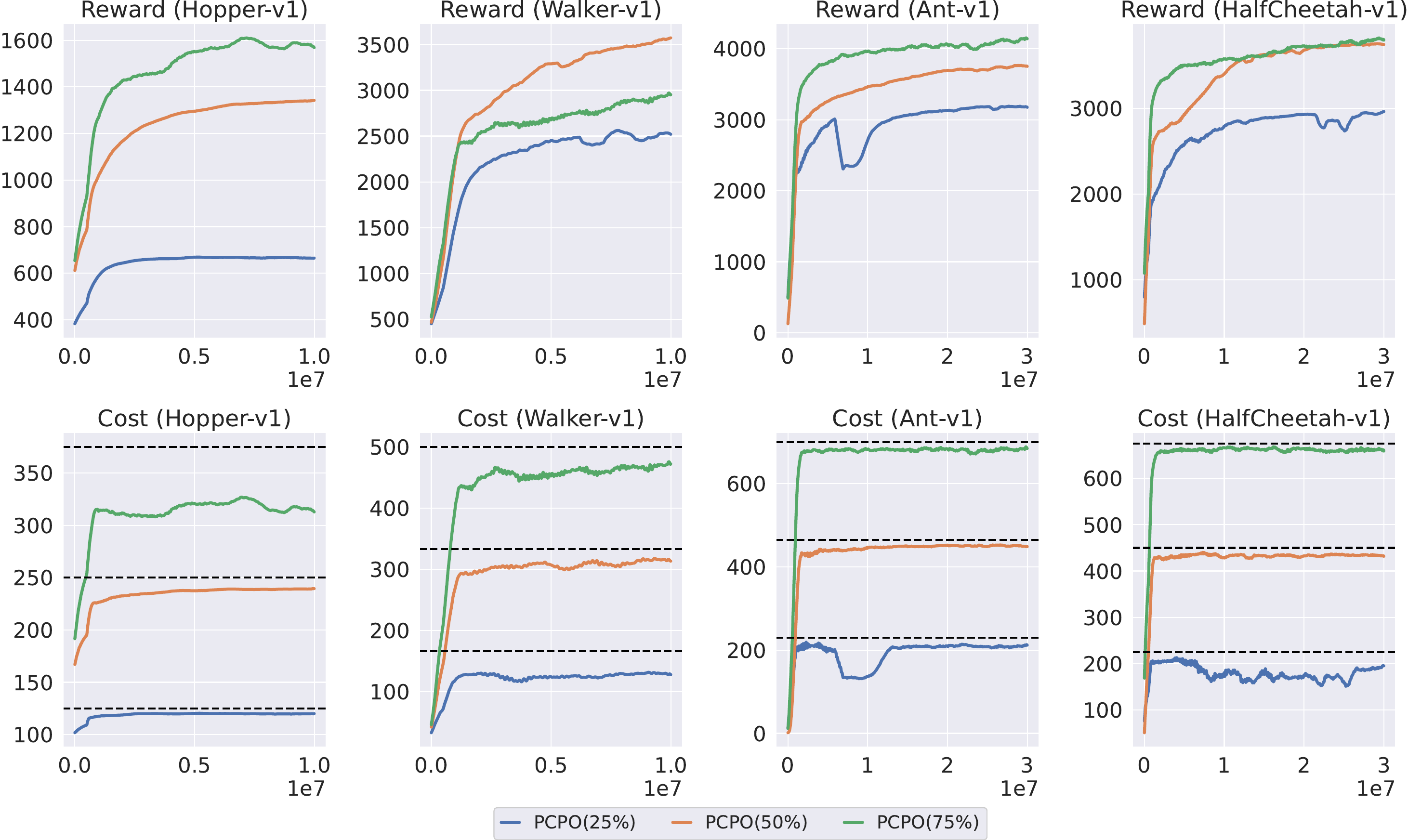}}
    \caption{The average performance of PCPO with different cost thresholds (about 1000 bootstrap samples over 6 random seeds).The x-axis indicates the total number of samples, while the y-axis shows the mean total reward/cost return from the most recent 100 episodes. Each solid line represents the PCPO trained with the corresponding threshold, illustrating the 25\%, 50\%, and 75\% cost requirements for the unconstrained TRPO agent. The shaded areas indicate the 95$\%$ confidence interval. }
    \label{figvc}
    \end{center}
\end{minipage}
\end{figure}

By training an unconstrained TRPO agent across different robotic models with velocity limitations, we generate 10 million samples to establish an initial cost baseline. Based on this baseline, we set cost thresholds at 25\%, 50\%, and 75\%, and subsequently train the PCPO algorithm.
Figure \ref{figvc} demonstrates that PCPO is able to learn policies that satisfy constraints under different cost thresholds in all test environments. Notably, even under stringent conditions with a cost threshold reduced to 25\%, such as in the Ant-v1 environment, both the reward and cost curves exhibit stability, indicating that the algorithm successfully adheres to cost constraints while maintaining high-performance levels. These findings not only confirm the effectiveness of PCPO at various cost thresholds but also demonstrate its capability to handle complex constraints across multiple robotic platforms.

\subsection{Generalization Analysis}\label{ap10}
To better assess the generalization ability of RL algorithms in practical applications, a strategy different from traditional supervised learning methods is adopted. Initially, the agent is trained on fixed random seeds. Subsequently, it is tested on 10 unseen random seeds. This method allows for evaluating algorithm performance under various unknown conditions, closely simulating real-world challenges.
Finally, all five algorithms are used to perform speed-limit tasks on robots to test their generalization ability. Table \ref{table2} presents the end results of trained agents on tasks involving speed limits. The experiments demonstrate that PCPO exhibits superior performance over most competing algorithms in the test environment, consistently achieving performance enhancement while maintaining lower costs. This achievement indicates that PCPO can effectively generalize to various test environments and consistently provide performance improvements across diverse tasks.

\subsection{Sensitive Analysis}\label{ap11}
\begin{figure}[htbp]
    \begin{center}    
    \centering{\includegraphics[width=1\linewidth, height=0.5\linewidth]{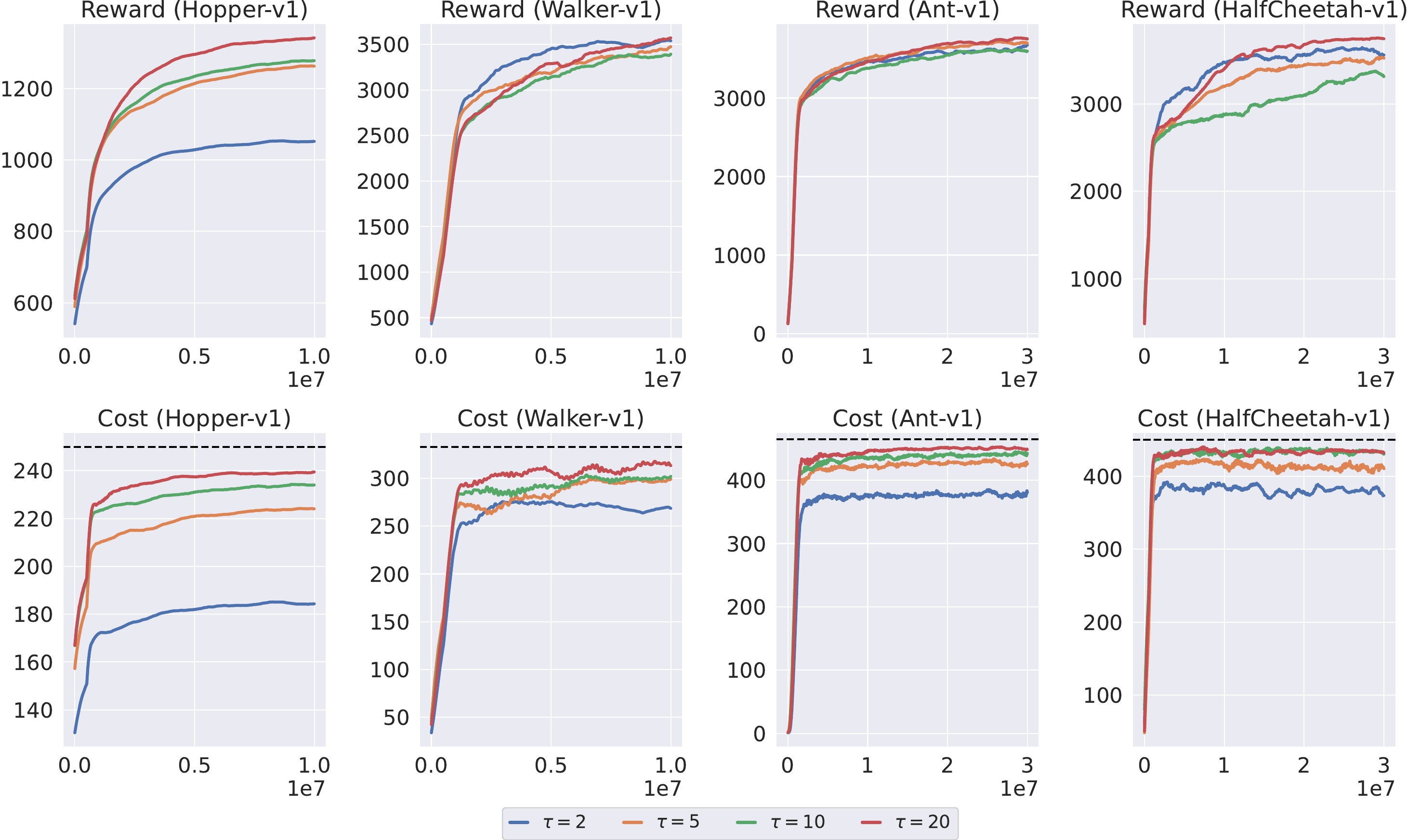}}
    \caption{Performance with respect to penalty factor $\tau$.}
    \label{figsensitive}
    \end{center}
\end{figure}

\begin{figure}[htbp]
    \begin{center}    
    \centering{\includegraphics[width=1\linewidth, height=0.5\linewidth]{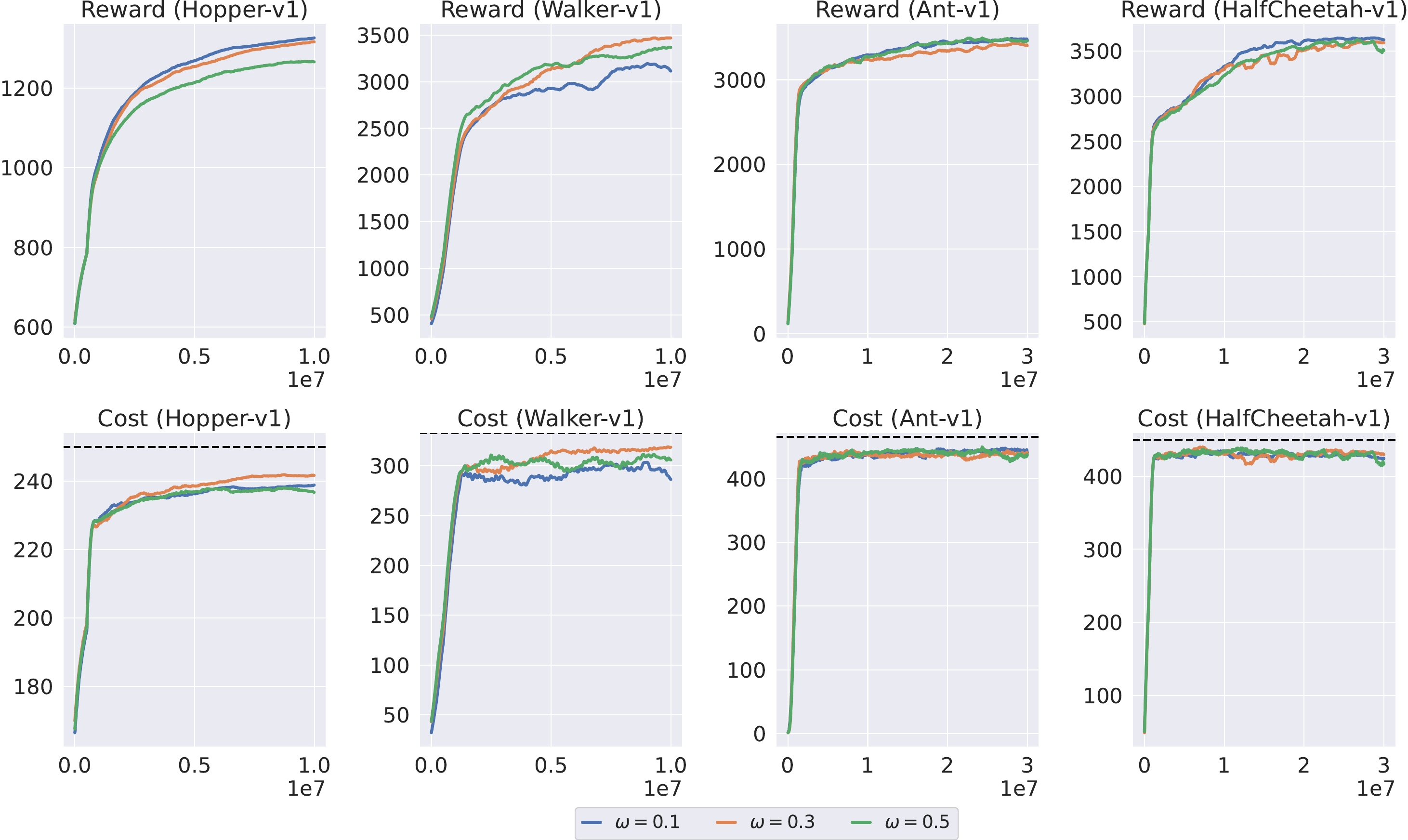}}
    \caption{Performance with respect to $\omega$.}
    \label{figsensitive2}
    \end{center}
\end{figure}

We test PCPO across 4 different values of $\tau$ while maintaining all other parameters fixed. For each of the robots involved in the speed limit experiment, we run PCPO with 10 million samples on Hopper-v1, Walker-v1 with 30 million samples on Ant-v1, and HalfCheetah-v1. The results are shown in Figure \ref{figsensitive}. Our analysis indicates that as $\tau$ increases, both the reward and cost values exhibit a corresponding upward trend. Nonetheless, our method is capable of achieving the desired reward, even with the lowest value of $\tau$, which underscores the robustness of our approach to variations in parameters. In Figure \ref{figsensitive2}, we visualize the influence of $\omega$ in the speed limit experiments. We observe that the performance of PCPO remains robust across different values of this parameter. 

\begin{table}[htbp] 
    \caption{\label{tabcircle}The mean and 95$\%$ confidence interval, derived from 1000 bootstrap samples across 6 random seeds, illustrate the reward/cost return following training in circle environments. Cost thresholds are noted in brackets beneath the environment names.}
     \setlength{\tabcolsep}{2pt}
    \begin{center}
    \begin{small}
    \begin{sc}    
    \begin{tabular}{cccccccc}
        \toprule
        Environment & & CUP & EPO & FOCOPS & TRPOLag & PCPO \\
        \hline
        PointCircle1-v0 & Return & $44.58\pm3.94$ & $45.67\pm1.55$ & $47.03\pm3.41$  & $48.3\pm3.03$ & $\bm{49.49\pm0.36}$ \\
        (100.) & Cost & $76.63\pm36.79$ & $95.65\pm5.66$ & $88.51\pm45.65$ & $101.8\pm19.97$ & $88.72\pm4.27$\\
        \hline
        CarCircle1-v0 & Return & $18.89\pm0.54$ & $16,43\pm0.58$& $19.85\pm0.13$ & $18.99\pm0.75$ & $\bm{20.07\pm0.26}$ \\
        (100.) & Cost & $99.86\pm16.89$ & $87.66\pm12.35$ & $98.98\pm5.66$ & $107.74\pm9.72$ & $88.77\pm1.74$\\
        \hline
        \bottomrule
    \end{tabular}
    \end{sc}
    \end{small}    
    \end{center}
\end{table} 

\subsection{Ablation Study}\label{ap12}
\begin{figure}[htbp]
    \begin{center}    
    \centering{\includegraphics[width=1\linewidth, height=0.5\linewidth]{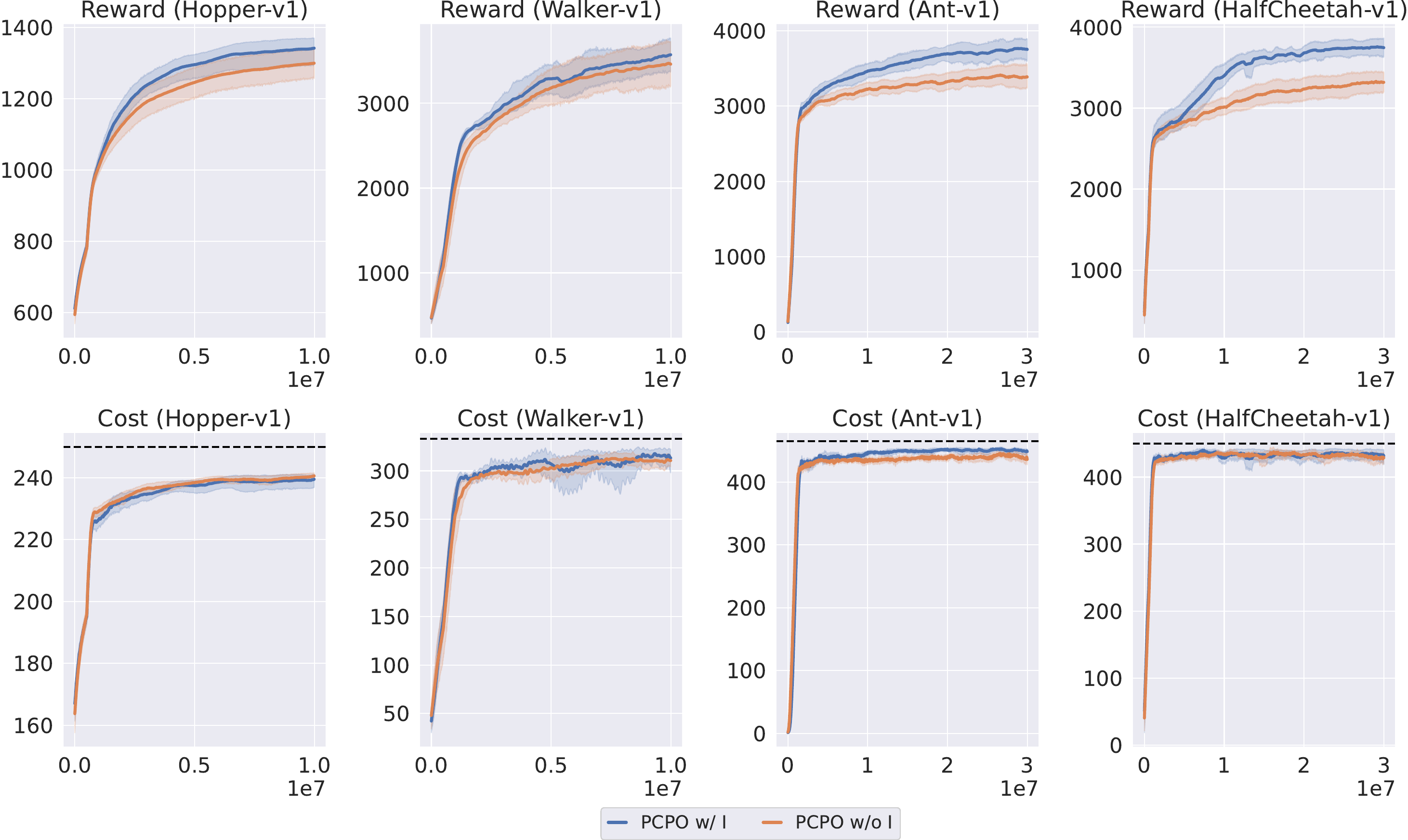}}
    \caption{PCPO ablation study. The average performance (about 1000 bootstrap samples over 6 random seeds). The x-axis indicates the total number of samples, while the y-axis shows the mean total reward/cost return from the most recent 100 episodes. The shaded areas indicate the 95\% confidence interval.}
    \label{figablation}
    \end{center}
\end{figure}

\begin{figure}[htbp]
\begin{center}    
    \centering{\includegraphics[width=1\linewidth, height=0.5\linewidth]{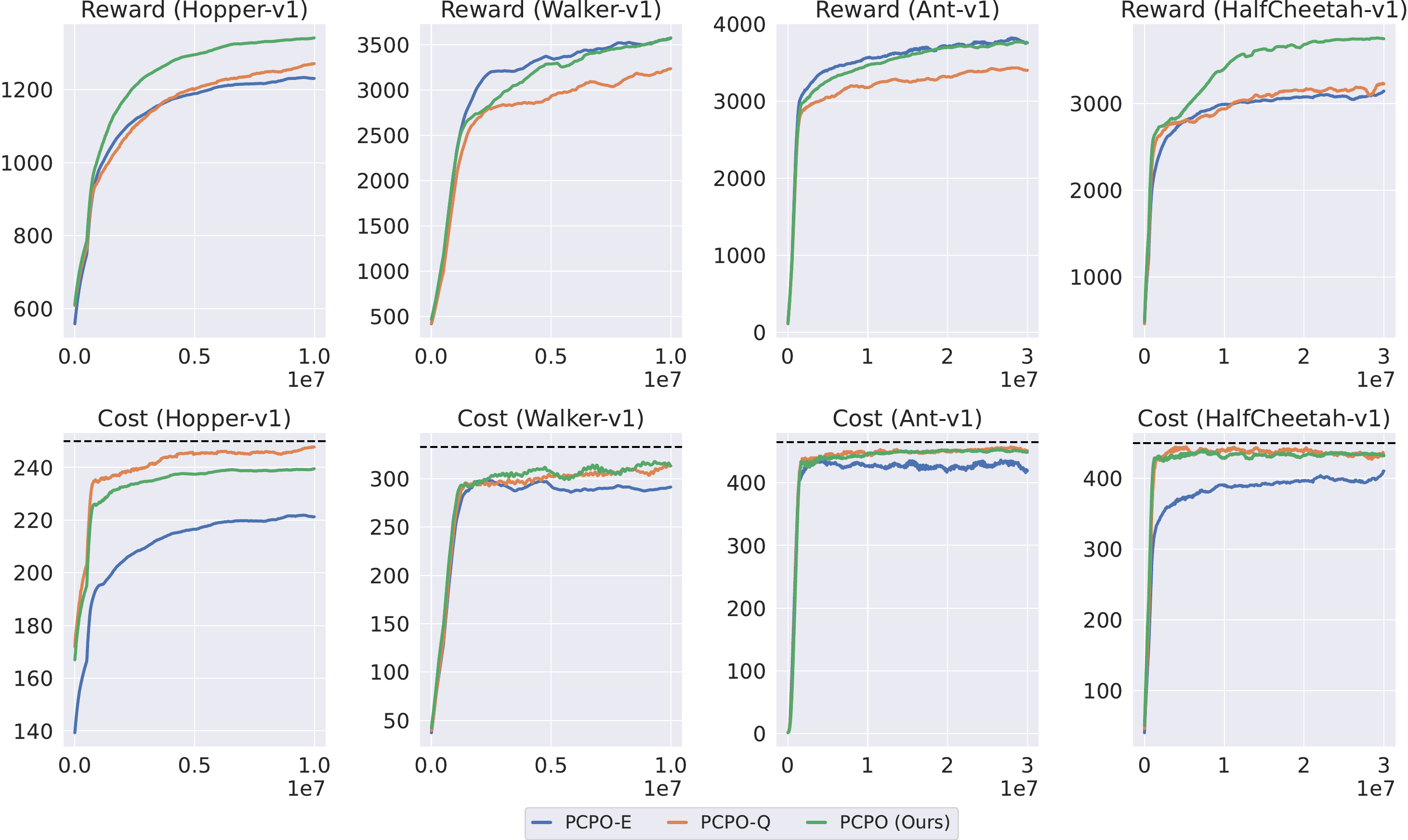}}
    \caption{Performance with different barrier functions.}
    \label{figbarrier}
    \end{center}
\end{figure}

We execute this experiment on the safe velocity tasks. We ablate the model's performance without intrinsic reward in Figure \ref{figablation}. For the effectiveness of our designed barrier function, we compared it with the quadratic (PCPO-Q) and exponential barrier functions (PCPO-E) in the Safe Velocity environment. The quadratic barrier function is defined as $\varphi_\tau^q = \tau g_{C_i}(\pi_\theta)^2$, while the exponential barrier function is given by $\varphi_\tau^e = e^{\tau g_{C_i}(\pi_\theta)}$, and we set $\tau = 1$ and $\tau = 0.01$ for them respectively which we found works well in our experiments. Since the inverse barrier function, when combined with our method, yielded poor performance, we omit its results for clarity. To ensure a fair comparison, we kept all other components of our algorithm unchanged across different barrier functions. The results, presented in Figure \ref{figbarrier}, demonstrate that our method achieves the most stable performance and consistently attains near-optimal rewards across all four robotic types.  

\section{Pseudocode}\label{Pseudocode}
The following pseudocode (Algorithm 2) details the complete implementation workflow of the PCPO algorithm, which integrates the preemptive penalty mechanism and constraint-aware intrinsic reward as core components. This step-by-step procedure formalizes how PCPO achieves safe policy optimization by balancing reward maximization, constraint satisfaction, and boundary-aware exploration.

\begin{algorithm}[htbp]
\caption{PCPO Algorithm}
\begin{algorithmic}[1]
\State Initialize: Value networks for reward $V_{\phi}$ and costs $V_{\psi}^{C_{i}}$, Policy network $\pi_\theta$.
\State Initialize: Discount rate $\gamma$, GAE parameter $\beta$; Logarithmic barrier function parameter $\tau$; Learning rates $\alpha_r$; Trust region bound $\delta$; Cost bound $d_i$.
\While{stopping conditions have not been satisfied}
    \State Collect batch data of $M$ episodes each of length $T$, containing tuples $(s_{i,t}, a_{i,t}, R_{i,t}, s_{i, t+1}, C_{i, t})$ sampled using policy $\pi_\theta$, for $i=1,...,M$ and $t=1,...,T$.
    \State Compute the average C-return across all episodes:
 $$\hat{J}_C = \frac{1}{M}\sum_{i=1}^M\sum_{t=0}^{T-1}\gamma^t C_{i, t}$$  
    \State Estimate advantage functions $\hat{A}_{i, t}$ and $\hat{A}_{i,t}^C$, $i=1,...,M,t=1,...,T$ using GAE.
    \State Estimate intrinsic reward $I_{C_i}, i=1,...M$.
    
    \State Get $V_{i,t}^{target}=\hat{A}_{i,t} + V_\phi(s_{i,t})$ and $V_{i,t}^{C, target} = \hat{A}_{i,t}^C + V_\phi^C(s_i,t)$
    \State Estimate $\nabla \hat G(\theta ), \nabla {\hat \varphi _\tau }({g_C}_{_i}(\theta ))\nabla {g_C}_{_i}(\theta ), \nabla I_{C_i}^{\pi_\theta}(\theta), \hat H$ with size $B$
    \State Obtain $\pi_\theta$ by backtracking line search to enforce satisfaction of sample estimate of ${D_{KL}}(\pi_\theta||{\pi _{{\theta _k}}}) \le \delta$
    \For{epoch $j = 1,2,...$}
        \For{each minibatch $$\left\{s_j, a_j, A_j, A_j^C, V_j^{target}, V_j^{C, target}\right\}$$ of size $B$}
            \State Value loss functions
            \State $\mathcal{L}_V(\phi) = \frac{1}{2B}\sum_{j=1}^B(V_\phi(s_j) - V_j^{target})^2$  
            \State $\mathcal{L}_{V^C}(\psi) = \frac{1}{2B}\sum_{j=1}^B(V_\psi(s_j) - V_j^{C, target})^2$  
            \State Update value networks
                \State $\phi \leftarrow \phi - \alpha_r \nabla_\phi \mathcal{L}_V(\phi),\psi \leftarrow \psi - \alpha_r \nabla_\phi \mathcal{L}_{V^C}(\psi)$ 
        \EndFor
    \EndFor
\EndWhile
\end{algorithmic}
\end{algorithm}



\end{document}